\documentclass[12pt]{spieman}
\usepackage{amsmath,amsfonts,amssymb}
\usepackage{graphicx}
\usepackage{setspace}
\usepackage{tocloft}
\graphicspath{{figures/}}
\usepackage{xcolor}
\usepackage{soul}
\usepackage{ulem}
\usepackage{booktabs}
\usepackage[export]{adjustbox}
\usepackage{makecell}
\usepackage[firstpage=true,color=black,placement=top,scale=1,opacity=1,vshift=-1cm]{background}

\SetBgContents{
\begin{minipage}{\linewidth}
\small
Copyright 2017 Society of Photo Optical Instrumentation Engineers (SPIE).\\
Journal of Medical Imaging 4(4), 044002 (27 October 2017).\\
\url{http://dx.doi.org/10.1117/1.JMI.4.4.044002}
\end{minipage}
}

\title{Mesh-to-raster region-of-interest-based non-rigid registration of multi-modal images}

\author[a,*]{Rosalia Tatano}
\author[a]{Benjamin Berkels}
\author[b]{Thomas M. Deserno}

\affil[a]{Aachen Institute for Advanced Study in Computational Engineering Science (AICES), RWTH Aachen University, Schinkelstra\ss e 2, 52062 Aachen, Germany}
\affil[b]{Peter L. Reichertz Institute for Medical Informatics, University of Braunschweig -- Institute of Technology and Hannover Medical School, M\"uhlenpfordtstr. 23, 38106 Braunschweig, Germany}

\cftpagenumbersoff{figure}
\cftpagenumbersoff{table}
\begin{document}
\maketitle

\begin{abstract}
Region of interest (ROI) alignment in medical images plays a crucial role in diagnostics, procedure planning, treatment, and follow-up. Frequently, a model is represented as triangulated mesh while the patient data is provided from CAT scanners as pixel or voxel data. Previously, we presented a 2D method for curve-to-pixel registration. This paper contributes (i) a general mesh-to-raster (M2R) framework to register ROIs in multi-modal images; (ii) a 3D surface-to-voxel application, and (iii) a comprehensive quantitative evaluation in 2D using ground truth provided by the simultaneous truth and performance level estimation (STAPLE) method. The registration is formulated as a minimization problem where the objective consists of a data term, which involves the signed distance function of the ROI from the reference image, and a higher order elastic regularizer for the deformation. The evaluation is based on quantitative light-induced fluoroscopy (QLF) and digital photography (DP) of decalcified teeth. STAPLE is computed on 150 image pairs from 32 subjects, each showing one corresponding tooth in both modalities. The ROI in each image is manually marked by three experts (900 curves in total). In the QLF-DP setting, our approach significantly outperforms the mutual information-based registration algorithm implemented with the Insight Segmentation and Registration Toolkit (ITK) and Elastix.
\end{abstract}


{\noindent \footnotesize\textbf{*}Rosalia Tatano,  \linkable{tatano@aices.rwth-aachen.de} }

\vspace{1cm}

\section{Introduction}
\label{sect:intro}
Two-dimensional (2D) as well as three-dimensional (3D) images play a crucial role in diagnosis, treatment planning, treatment, and the assessment of progression and/or regression of a condition or disease in a patient. In this context, comparisons of subject data acquired using different imaging modalities or of subject and model data are often necessary. Hence, multi-modal image registration methods aim at aligning images (2D) or volumes (3D) acquired with different devices, thus integrating the information provided by this data. The goal is to find the optimal transform that best aligns structures in two input 2D or 3D images \cite{Keszei16}.

Mutual information (MI) has been widely used in multi-modal image registration \cite{viola1997, pluim2003}. The idea is to maximize iteratively the MI between the two images globally with respect to the transformation. This is equivalent to minimizing the joint entropy of the two data sets, which occurs when the two images or the two volumes are correctly registered.

However, in several cases, there is a particular region of interest (ROI) predefined in the medical recording, such as a tumorous region, a lesion, or some other pathologies. Therefore, accurate ROI alignment is of primary importance. Moreover, in some applications, the ROI might change, e.g., due to pathology or tumor growth. Using MI to register these ROIs might lead to inaccuracies, since MI is computed on the entire image disregarding the corresponding and overlapping parts of the images and hence, is sensitive to the ROI's size and content \cite{pluim2003}.

Therefore, ROI-based registration has been addressed in several works already. Wilkie et al.\cite{wilkie2005}, proposed a modification of MI registration that takes into account information from the ROIs, using a weighted combination of the statistics of the entire images and of the ROIs. However, the registration is affected by the value of the weighting parameter, which is difficult to determine, and the ROIs probability distributions. Yavariabdi et al. presented a registration method for magnetic resonance imaging (MRI) and transvaginal ultrasound (TUS) that matches manually marked contours of the ROIs in the two modalities through a one-step deformable iterative closest point (ICP) method\cite{yavariabdi2013}. Gu et al.\cite{gu2013} have proposed a contour-guided deformable image registration (DIR) algorithm for adaptive radiotherapy that deforms images using a demons registration algorithm with an additional regularization term based on modified image intensities inside manually marked ROIs. These modified images differ from the original image only if the intensities of the ROIs differs from the surrounding. Finite element model (FEM) -based deformable registration has been employed in Penjweini et al.\cite{Penjweini2016} to match the contours of the ROIs in a series of computed tomography (CT) scans of the lungs acquired pre-operatively with intra-operative images acquired using an infrared camera-based navigation system during the surgery stage of pleural photodynamic therapy. Zhong et al. \cite{zhong2016} proposed a method for a dental implant guide that uses ICP on teeth contours point sets extracted from CT scans and from 3D patient model's cross sections to retrieve the rigid transformation between two images. A registration method for dynamic contrast-enhanced MRI (DCE-MRI) images of the liver is proposed by Zhang et al.,\cite{zhang2016} but the energy to be minimized is evaluated just on the segmented liver, disregarding the other image areas completely.

Whenever considering a ROI, automatic segmentation of such is important. Obviously, the accuracy of the registration depends on the accuracy of segmentation. ROI segmentation is frequently based on the pixel or contour levels, and several approaches have been proposed in the literature.  By way of example, the following approaches are taken from dentistry, where digital photography (DP) and quantitative light-induced fluoroscopy (QLF) are common 2D imaging modalities:
\begin{itemize}
\item { manual draw}: Hope et al. \cite{hope2014} manually draw the contour  around the boundary of the tooth;
\item { gray scale}: A threshold technique based on the intensities of the tooth region and of the background is used in Yan et al. \cite{yan2011} to determine the tooth region in fluorescence and white light images. The underlying assumption is that the tooth region has higher intensities than the background;
\item { color}: A color based segmentation technique combined with morphological operations is used in Datta et al. \cite{datta2015} to segment a tooth from gum, lips and neighbors teeth in RGB images;
\item { statistical model}:  In Mansoor et al. \cite{mansoor2015}, an initial segmentation of the tooth was achieved using a Gauss-Markov random field statistical model and then refined by the practitioner;
\item { active contour}: In Shah et al. \cite{shah2006}, the contour of teeth from dental images is extracted using an active contour algorithm that depends on the intensity of the tooth region. However, the algorithm is sensitive to poor contrast in image intensities and the presence of neighboring teeth;
\end{itemize}

Another issue in registration is the multi-modality. Usually, multi-modal means that data from the same subject is taken with different imaging modalities, such as, for example, computed tomography (CT) and magnetic resonance imaging (MRI). Furthermore, atlas registration is required frequently in modern medicine to transfer knowledge coded in a general model (i.e., the atlas) to a specific subject (e.g., the patient in a diagnostic or therapeutic process). Such comparison of virtual physiological human (VPH) models\cite{viceconti2016} with subject-specific scan data bears another challenge for registration: the VPH models are usually in triangulated mesh-based coding, while patient measurements are obtained with computed axial tomography (CAT) scanners and stored as pixel or voxel data \cite{oliveira2016}. This yields curve-to-pixel and surface-to-voxel registration problems in 2D and 3D, respectively, disregarding whether registration is considered as global or local (i.e., ROI-based) problem.

In our previous work, a curve-to-pixel-based registration method has been presented \cite{berkelsSPIE2016, berkelsBVM2016} and used to align the ROIs of 2D images acquired with QLF and DP. The ROIs are segmented using a color space transform into grayscale, which were adapted to the imaging modality, and thresholded.
Registration is based on aligning the ROI's contours, i.e., the tooth region.
Our curve-to-pixel method allows superimposing DP with QLF and thus, a direct comparison of the detected demineralized areas,  an undesirable side effect of orthodontic treatment with fixed appliances \cite{Gorelick82}.
To the best of our knowledge, registration of QLF and DP for direct comparison of these two image modalities had not been explicitly addressed before.

In our study on the comparison of the demineralized areas in DP and QLF images of a tooth, the demineralized area on the tooth surface often is more evident in one modality than in the other, or is even indistinguishable in one of the modalities. Thus, the correlation between the demineralized areas shown in the two modalities needs to be investigated. Relying on ROI-features can induce a correlation bias in the registration results and therefore, using a registration method that does not use this information is desirable. In addition, while it is relatively easy to segment the tooth region in QLF images, this segmentation is more challenging in the photo due to the poor contrast between tooth and gum colors and the lack of separation of neighboring teeth.

Motivated by the setting outlined above, the aim of the present work is to provide a general registration methodology that is applicable in 2D and 3D, copes with different imaging modalities as well as types of data representations (mesh and raster-scan), and handles global and local (ROI-based) problems. The key features of the proposed method, whose combination sets it apart from the existing methods, are as follows:
\begin{itemize}
\item[i.] a contour-to-image approach: the proposed method is able to directly link data types of different dimension, e.g. it can align a surface to a volumetric image. Most existing methods can only align data of the same dimension and type, e.g., surface-to-surface or image-to-image;
\item[ii.] only ROI shape required: previous works rely on information about the interior of the entire image or the ROI (e.g., internal features, intensities, statistics...). The proposed method does not make any assumption on the features inside the ROIs. This is particularly suitable for studies where the correlation of the modalities inside the ROIs is unknown and supposed to be determined. In fact, the use of feature-based registration algorithms is bound to induce a bias towards the correlation assumed by the registration;
\item[iii.] no strict requirement on the ROI segmentation in the reference image: unlike previous works, the proposed method only requires an accurate contour of the ROI in the template. The classification of the reference image ROI only needs to provide sufficient information about the ROI's shape (see below for an example);
\item[vi.] topology preserving ROI segmentation: the method can also be used to segment the ROI in the reference image. In this case, the contour of the template image serves as initial guess for the segmentation of the ROI in the other image modality. This is helpful in case the reference image modality is difficult to segment but the topology of the segment is known;
\item[v.] nonlinear least-square problem: the proposed method leads to a nonlinear least-square problem, which can be solved efficiently using the Gauss-Newton algorithm. Many multi-modal registration approached like Mutual Information lead to much more involved optimization problems with fewer (or no) guarantees on optimality of minimization algorithms.
\end{itemize}
Due to the properties listed above, the proposed method is particularly suited for the QLF/DP registration problem the method was originally designed for. Indeed, the proposed method allows to align the tooth areas in the two modalities without relying on ROI-features (cf. ii). Due to the challenges posed by the photo classification into tooth area and background, neighboring teeth are still present. However, due to the fact that the bottom and the upper part the tooth region is clearly delineated, the proposed method is able to extract the contour of a target tooth (cf. iii), disregarding the neighboring teeth on the side of it, and align the corresponding tooth region to the one shown in the QLF, since the algorithm preserves the ROI topology (cf. i, iv). Nevertheless,
the method can be applied to a much larger class of registration problems.

Furthermore, we aim to comprehensively and reliably evaluate the general registration approach using sufficiently large and reliable data sets.

\section{Materials and methods}

In this section, we present a general mesh-to-raster (M2R) registration framework, its application, and the methodology used for quantitative evaluation.

\subsection{M2R registration method}

The M2R method is constructed in the continuous setting as minimization problem and then discretized. After the discretization, we propose a numerical minimization strategy. Moreover, a parametric registration is performed to provide a first alignment of the two input images.

\subsubsection{Continuous approach}

\label{subsec:MatMed}
Let us assume that we are given two data sets, named $f$ and $g$, of the same anatomical structure acquired with different modalities and that the image $f$ is given as mapping $f:\Omega\to\mathbb{R}^n$, $n\in\{2,3\}$, on the unit cube $\Omega=[0,1]^n$. Let $\mathcal{C}$ indicate a hypersurface, i.e. a curve for $n=2$ and a surface for $n=3$, representing the contour of the region of interest (ROI), which needs to be extracted from $g$, in case $g$ is also an image or may be identical to $g$, if $g$ was directly acquired as hypersurface, for instance, by a laser scan.

We denote the region of interest in the image $f$ by $S_f\subset\Omega$ and now want to find a non-rigid deformation $\phi:\Omega\to\mathbb{R}^n$ that matches $\mathcal{C}$ to the boundary of the set $S_f\subset\Omega$. To this end, let $d$ be the signed distance function of $S_f$, i.e. $d(c,S_f)=\pm \text{dist}(c,\partial S_f)$. Thus, $d$ is the Euclidean distance of the point $c$ to the boundary of $S_f$, where the sign is positive if $c$ is outside of $S_f$ and negative otherwise \cite{bauer}. Then, the desired alignment of $S_f$ and $\mathcal{C}$ is attained by minimizing the energy
\begin{equation}E[\phi] =  E_\text{match}[\phi] + E_\text{reg}[\phi] =  \frac{1}{2}\int_{\mathcal{C}}w_c\left(d(\phi(c), S_f)\right)^2 \mathrm{d}\mathcal{H}^{n-1}(c)+ \frac{\lambda^2}{2}\int_\Omega \Vert \Delta \phi(x)\Vert^2\mathrm{d}x,\label{eq:En}\end{equation}
where $\Delta \phi = (\Delta\phi_1, \dots, \Delta\phi_n)$ is the vector of the Laplacian of the components of $\phi$ and $\mathrm{d}\mathcal{H}^{n-1}$ denotes the $n-1$ dimensional Hausdorff measure. Thus, the first term is a hypersurface integral. In particular, it is a curve integral in case $n=2$ and a surface integral in case $n=3$.

The energy measures the distance of the deformed hypersurface to the boundary of the ROI in the image $f$ and the smoothness of the deformation.
The parameter $\lambda>0$ controls the smoothness of $\phi$. Here, $w_c>0$ are application dependent weights defined to control the influence of the hypersurface points. Since the data term is using the integral over the hypersurface, it involves only the deformation of $\mathcal{C}$. However, the use of a higher order regularizer extends the non-rigid deformation to the whole domain $\Omega$.

In contrast to our previous implementation \cite{berkelsSPIE2016, berkelsBVM2016}, here the data term is formulated as the integral over the hypersurface. Previously, the data term was defined using the sum over the hypersurface points of the weighted signed distance function calculated at the deformed hypersurface points. However, this approach might lead to problems if the hypersurface points are not approximatively equidistant. Using the integral instead, the distance of a hypersurface point with respect to its neighbors is taken into account.

\subsubsection{Discretization}

The deformation $\phi$ is expressed as displacement $u:\Omega\to\mathbb{R}^n$ via $\phi(x)=x+u(x)$, for $x\in\Omega$, noting that $\Delta\phi=\Delta u$.
For the spatial discretization of $u$, we use multilinear Finite Elements on a uniform rectangular grid on the image domain $\Omega$ \cite{braess}. Let $\{\psi_j\}_{j\in J}$ denote the FE basis functions, with nodal index set $J$. Let $M$ and $L$ denote the lumped mass matrix and the stiffness matrix respectively, i.e.
$M_{i,j}=\int_{\Omega}\mathcal{I}(\psi_i\psi_j)\mathrm{d}x\text{ and }L_{i,j}=\int_\Omega \nabla\psi_i\nabla\psi_j\mathrm{d}x,$
where $\mathcal{I}$ is the bilinear Lagrangian interpolation.
Although the chosen regularizer involves second derivatives, it can be approximated as $E_\text{reg}[u]=\frac{\lambda^2}{2}\sum_{i=1}^n\Vert M^{\frac{1}{2}}LU_i\Vert^2$ \cite{berkels}, using the lumped mass and stiffness matrices. Here, $U_i$ denotes the vector of nodal values that uniquely represents the scalar FE function $u_i$.

\subsubsection{Minimization}
\label{sec:GenMinimization}

In contrast to our previous implementation \cite{berkelsBVM2016}, the data term now is evaluated using simplicial finite elements on the hypersurface using a quadrature rule for the numerical evaluation of the integral. The minimization of $E$ is formulated as nonlinear least squares problem,  $E[u]=\frac{1}{2}\Vert F[u] \Vert^2$, where
\begin{equation}\label{eq:VecFGeneral}F[u]= \left[\left\{\sqrt{w_c}\sqrt{m_k^q}d(x_k^q+u(x_k^q), S_f)\right\}_{\substack{k=1,\dots,N\\q=1,\dots,Q}}, \lambda M^{-\frac{1}{2}}LU_1, \dots,\lambda M^{-\frac{1}{2}}LU_n\right]^T.\end{equation} Here, $m_k^q$ are the weights corresponding to the $q$-th quadrature point $x_k^q$ in the $k$-th simplex describing the hypersurface. $N$
 and $Q$ denote the total number of simplices and of quadrature points in a simplex, respectively. The minimization is efficiently solved using the Gauss-Newton method \cite{Gratton07p106}.
To avoid the minimization from getting stuck in local minima, the non-rigid registration problem is solved for decreasing values of the parameter $\lambda$. For the sake of simplicity, we use this strategy to avoid a a multi level approach, which would also require to create a multi level representation of the unstructured simplicial grid of the hypersurface.

\subsection{Parametric registration algorithm}
To provide a reasonable initial guess for the Gauss-Newton algorithm above, a regularized parametric registration \cite{chum} is performed. The aim is to find an affine deformation $\varphi( c) = Ac+t $, where $A$ is a $n\times n$ matrix and $t$ a translation vector, that minimizes the energy $E[\varphi] =  E_\text{match}[\varphi] + E_\text{reg}^\text{par}[\varphi]$, with \begin{equation}
     \begin{aligned} E_\text{match}[\varphi] ={}& \frac{1}{2}\int_\mathcal{C}w_c\left(d(A c+t, S_f)\right)^2\mathrm{d}\mathcal{H}^{n-1}(c),\\  E_\text{reg}^\text{par}[\varphi]:={}&\frac{\alpha^2}{2}\int_\Omega \Vert J(\varphi(x)-x)\Vert_F^2\;\mathrm{d}x+ \frac{\mu^2}{2}\int_\Omega\sum_{i=1}^n \left\Vert I-\frac{1}{\partial_{x_i}(\varphi_i(x)-x_i)}D_{J(\varphi(x))}\right\Vert_F^2\mathrm{d}x\\={}&  \frac{\alpha^2}{2}\sum_{i=1}^n\int_\Omega \Vert \partial_{x_i}((A-I)x)\Vert^2\mathrm{d}x + \frac{\mu^2}{2}\sum_{i=1}^n \left\Vert I-\frac{1}{a_{ii}}D_A\right\Vert_F^2, \end{aligned}\label{eq:regPar}\end{equation}
\noindent
where $J(\varphi(x))$ indicates the Jacobian of $\varphi$ with respect to $x$ and $\varphi_i(x)$ represents the $i$-th component of the vector $\varphi(x)$. Here, the data term is the same as the non-rigid model. In contrast to our previous formulation \cite{berkelsBVM2016}, here as prior for the deformation the sum of two terms is used, with positive scalars $\alpha$ and $\mu$ that weight the contribution of these terms to the value of the energy.
The first term in the regularizer is the Dirichlet energy of the displacement. Noting that $J(\varphi(x))=A$, the second term is the squared Frobenius norm of the matrix $I-\frac{1}{a_{ii}}D_A$, where $I$ is the identity matrix and $D_A$ is a diagonal matrix whose entries are the diagonal entries of the matrix $A$.
Using this term, the difference of all the possible ratios of the diagonal entries of $A$ from 1 is penalized, thus isotropic scalings are preferred to anisotropic ones.

Similarly to the non-rigid case, the parametric registration is formulated as a least squares problem by defining the vector $F[\varphi]$ and solved using Gauss-Newton.

\subsection{Applications}

In this section, we demonstrate how to apply the method for solving registration problems in 2D and 3D.

\subsubsection{2D Example}
\label{sec:2D}
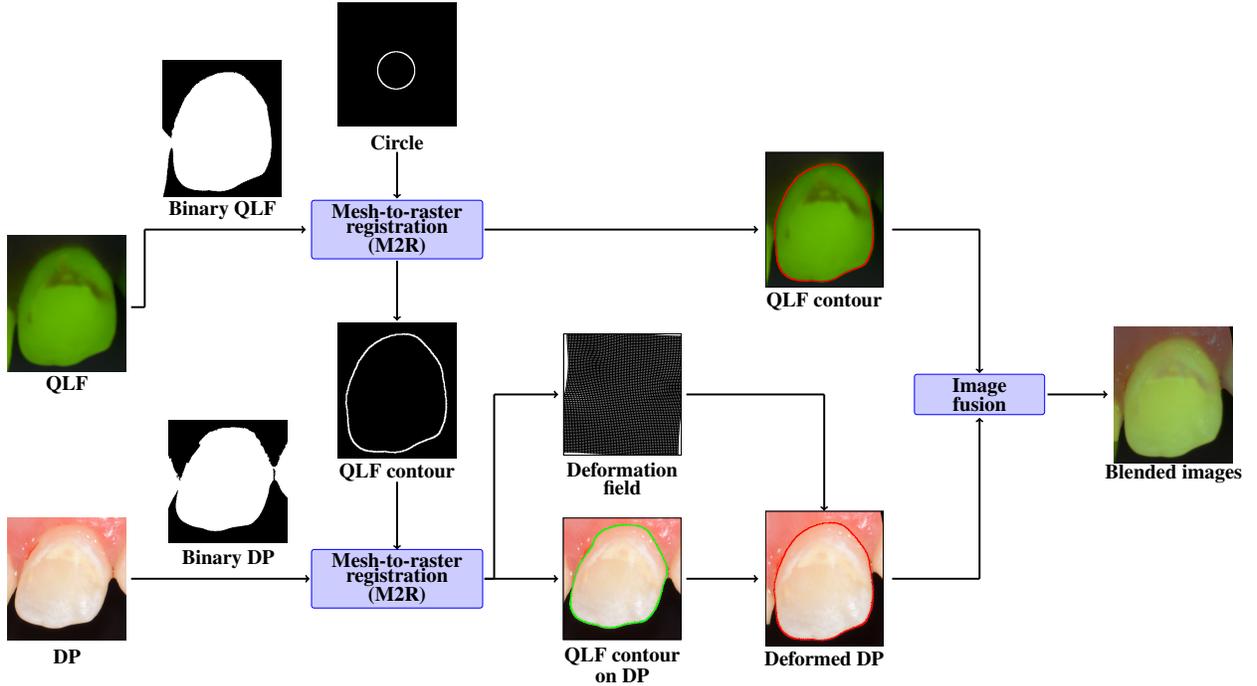
\begin{figure}[b]
\centering
\resizebox{\linewidth}{!}{
\begin{tikzpicture}
\node at (-17,2.2){\includegraphics[width=6.1cm]{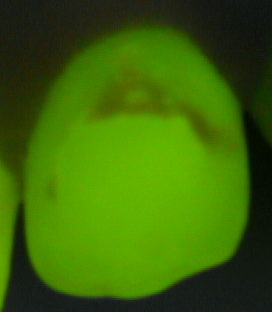}};
\node[font=\fontsize{30}{20}\selectfont] at (-17, -2){\bf QLF};
\node[draw, rectangle, align=center, inner sep=1ex, font=\sffamily, draw=blue, thick, fill=blue!20, text width=20em,  font=\fontsize{30}{20}\selectfont, rounded corners] at (0.,6) {\bf{Mesh-to-raster}\\ registration (M2R)};
\draw [line width=1mm, -] (-13.7,2) -- (-13,2);
\draw [line width=1mm, -] (-13,2.) -- (-13,6);
\draw [line width=1mm, ->] (-13,6) -- (-4.7,6);
\node at (-9,11.2){\includegraphics[width=6.1cm]{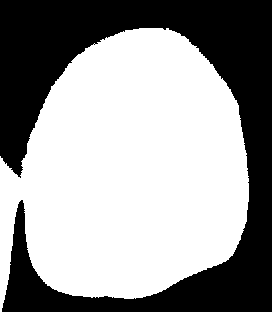}};
\node[font=\fontsize{30}{20}\selectfont,] at (-9, 7.){\bf Binary QLF};

\node at (0,14.5){\includegraphics[width=6.1cm]{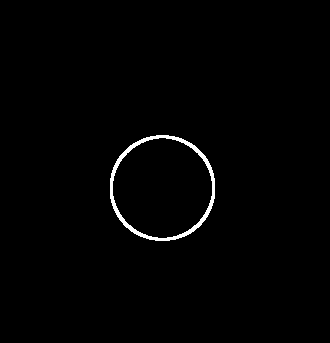}};
\node[font=\fontsize{30}{20}\selectfont,] at (0, 10.5){\bf Circle};
\draw [line width=1mm, ->] (0,10) -- (0.,7.6);

\node at (-17,-12){\includegraphics[width=6.1cm]{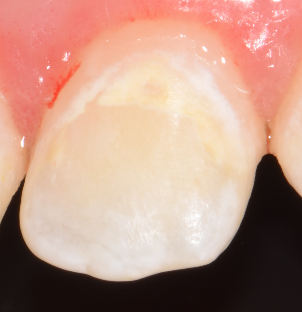}};
\node[font=\fontsize{30}{20}\selectfont] at (-17, -16){\bf DP};

\node[draw, rectangle, align=center, inner sep=1ex, font=\sffamily, draw=blue, thick, fill=blue!20, text width=20em,  font=\fontsize{30}{20}\selectfont, rounded corners] at (0.,-12) {\bf{Mesh-to-raster}\\ registration (M2R)};
\draw [line width=1mm, ->] (-13.7,-12) -- (-4.5,-12);

\node at (-8.7,-7){\includegraphics[width=6.1cm]{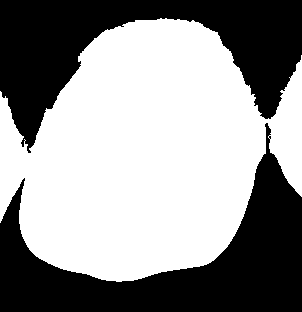}};
\node[font=\fontsize{30}{20}\selectfont,] at (-8.7, -11){\bf Binary DP};

\draw [line width=1mm, ->] (0,4.4) -- (0,1.3);
\draw [line width=1mm, ->] (0,-7) -- (0,-10.5);
\node at (0,-2.3){\includegraphics[width=6.1cm]{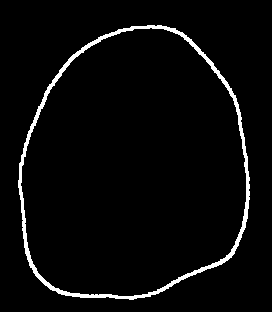}};
\node[font=\fontsize{30}{20}\selectfont,] at (0, -6.5){\bf QLF contour};

\draw [line width=1mm, ->] (5,-2.5) -- (8.3,-2.5);
\draw [line width=1mm, -] (5,-2.5) -- (5,-12);
\draw [line width=1mm, ->] (4.5,-12) -- (8.3,-12);
\draw [line width=1mm, -] (4.5,-12) -- (5,-12);

\node at (11.6,-2.5){\includegraphics[width=6.1cm]{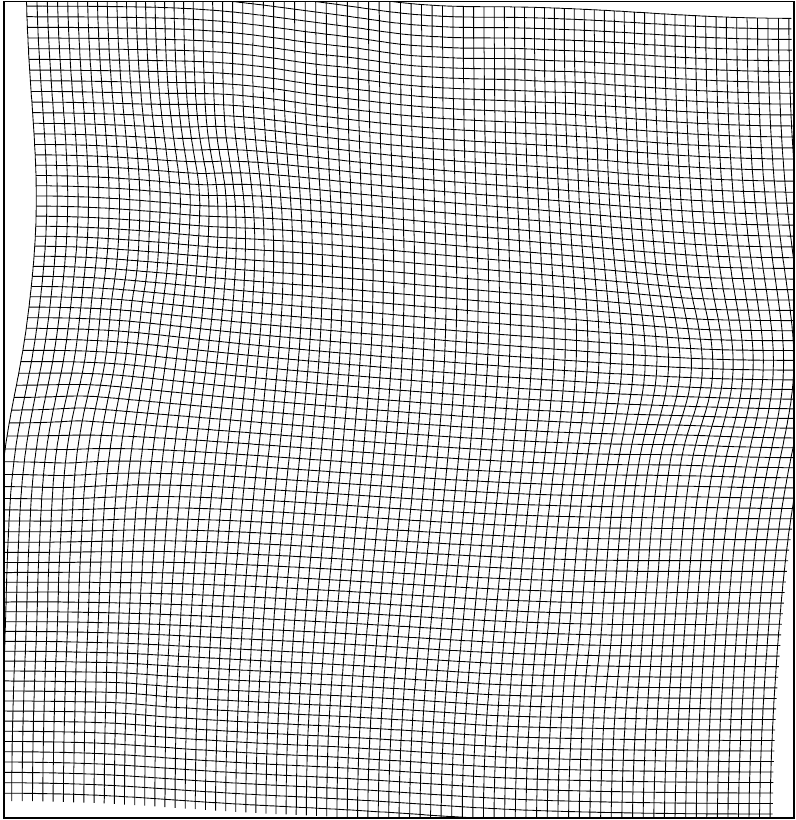}};
\node[font=\fontsize{30}{20}\selectfont] at (11.6, -6.4){\bf Deformation };
\node[font=\fontsize{30}{20}\selectfont] at (11.6, -7.3){\bf field };
\draw [line width=1mm, -] (14.9,-2.5) -- (22,-2.5);
\draw [line width=1mm, ->] (22,-2.5) -- (22,-8.5);

\node at (11.6,-12){\includegraphics[width=6.1cm]{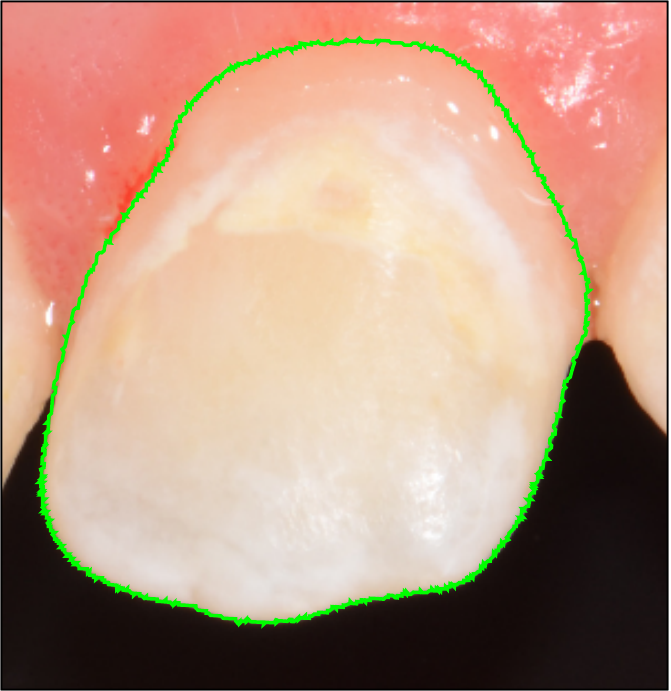}};
\node[font=\fontsize{30}{20}\selectfont] at (11.6, -16){\bf QLF contour};
\node[font=\fontsize{30}{20}\selectfont] at (11.6, -17){\bf on DP};

\node at (22,6.5){\includegraphics[width=6.1cm]{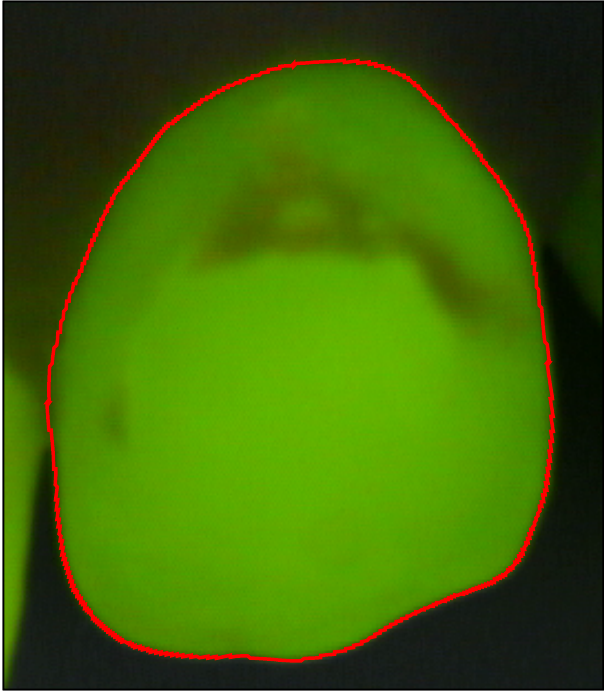}};
\node[font=\fontsize{30}{20}\selectfont,] at (22, 2.3){\bf QLF contour};
\draw [line width=1mm, ->] (4.5,6) -- (18.6,6);

\node at (22,-12){\includegraphics[width=6.1cm]{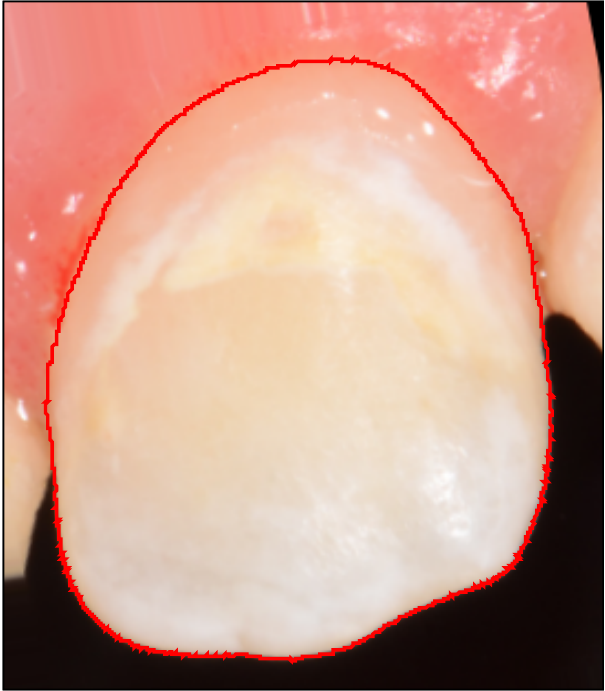}};
\node[font=\fontsize{30}{20}\selectfont] at (22, -16.1){\bf Deformed DP};
\draw [line width=1mm, ->] (14.9,-12) -- (18.6, -12);

\node[draw, rectangle, align=center, inner sep=1ex, font=\sffamily, draw=blue, thick, fill=blue!20, text width=15em,  font=\fontsize{30}{10}\selectfont, rounded corners] at (30.,-2.5) {\bf{Image}\\ fusion};
\draw [line width=1mm, ->] (33.5,-2.5) -- (36.6, -2.5);
\draw [line width=1mm, -] (25.5,6) -- (30, 6);
\draw [line width=1mm, ->] (30,6) -- (30, -1.4);
\draw [line width=1mm, -] (25.5,-12) -- (30, -12);
\draw [line width=1mm, ->] (30,-12) -- (30, -3.7);

\node at (40,-2.52){\includegraphics[width=6.1cm]{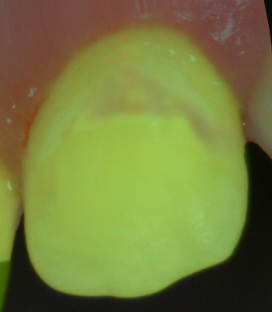}};
\node[font=\fontsize{30}{20}\selectfont] at (40, -6.6){\bf Blended images};
\end{tikzpicture}}
\caption{\label{fig:Method} Workflow of the curve-to-pixel registration method: a DP/QLF pair is given as input; first the algorithm is used to extract the contour of the tooth from the QLF image; then the segmented contour and the DP are used as input for the algorithm, obtaining the segmented DP and a deformation field that is used to align the DP to the QLF. Finally, an image that blends the QLF and the deformed DP can be created.}
\end{figure}

The selected 2D application aims at registering the tooth as ROI in DP and QLF for demineralization assessment. In contrast to the state of the art in ROI segmentation, we apply the proposed curve-to-image registration method for both the extraction of the tooth contour from the QLF (ROI segmentation) and for its alignment to the tooth region shown in the DP (Fig. \ref{fig:Method}).

\begin{itemize}

\item {\em Segmentation}:
In the QLF segmentation step, the tooth contour is represented as a small circle in the center of the image, since a tooth is topologically equivalent to a circle. The size of the circle was chosen to ensure that its entire boundary is within the tooth. Then, the registration algorithm aligns the circle to the boundary of the tooth region shown in the QLF, thus, yielding  the shape of the tooth\cite{berkelsBVM2016}. For the alignment of DP and QLF, the proposed algorithm is applied to the extracted QLF tooth contour and the tooth region shown in the DP.

\item {\em Registration}:
We assume that the curve is discretized as set of line segments between points $\{c_i\}_{i=1}^N$, where $N$ is the number of points. The data term is discretized using the midpoint rule over each line segment $[c_i,c_{i+1}]$ as
\begin{equation}E_\text{match}[\phi]= \frac{1}{2}\int_{\mathcal{C}}w_c\left(d(\phi(c), S_f)\right)^2 dc \approx \sum_{i=1}^Nw_{c_i}l_{c_i}\left(d\left(\phi\left(c_{i+\frac{1}{2}}\right), S_f\right)\right)^2\end{equation}
where $c_{i+\frac{1}{2}}$ indicates the midpoint of the line segment with endpoints $c_i$ and $c_{i+1}$ and $l_{c_i}$ its the length, i.e. $l_{c_i}=\Vert c_{i+1}-c_i\Vert$. The weights $w_{c_i}$ of the data term in Eq. (\ref{eq:En}) are used to get a proper alignment of the curve points on the vertical boundary of the tooth. In fact, these points may have no counterpart in the boundary of $S_f$, since often the thresholded images do not exhibit a clear separation between a tooth and its neighbors. Thus, the weights $w_{c_i}$ are defined using the vector $\vec{v}_{c_i}=\frac{1}{2}(c_{i+1}-c_i)$ that characterizes the orientation of the curve. The bigger the absolute value of the $x$-component of $\frac{\vec{v}_{c_i}}{|\vec{v}_{c_i}|}$, the less vertical is $C$ at $c_i$. Hence, $w_{c_i}$ it set to this value.

Thus, the vector $F$ \eqref{eq:VecFGeneral} that encodes the non-linear least squares problem in the 2D setting is
\begin{equation}F[u]= \left[\left\{\sqrt{w_{c_i}l_{c_i}}d\left(c_{i+\frac{1}{2}}+u\left(c_{i+\frac{1}{2}}\right), S_f\right)\right\}_{i=1,\dots,N}, \lambda M^{-\frac{1}{2}}LU_1,\lambda M^{-\frac{1}{2}}LU_2\right]^T.
\end{equation}
The vector $F$ for the parametric registration step is defined similarly (see Appendix \ref{app-1}).

At each iteration of the Gauss-Newton algorithm, the resulting linear system is solved using a sparse Cholesky factorization of the matrix $J_F^TJ_F$, where $J_F$ denotes the Jacobian of $F$. To this end, we use CHOLMOD from the SuiteSparse\cite{cholmod2008} library.

\end{itemize}

\subsubsection{3D Example}
\label{sec:3D}
The 3D use-case is taken from the Regional Anaesthesia Simulator and Assistance (RASimAs) project \cite{rasimas}, where subject-specific MRI need to be registered to a VPH model composed of mesh-based surfaces for skin, fascia, muscle, bone, vessels, and nerves.

Hence, we assume that $f$ is obtained from a volumetric scan, the surface $\mathcal{C}$ represents the ROI contour of $g$, and that $g$ is given as a triangle mesh. Then, the data term is discretized using linear FE on the triangle mesh as
\begin{equation}E_\text{match}[\phi]= \frac{1}{2}\int_{\mathcal{C}}w_c\left(d(\phi(c), S_f)\right)^2 dc \approx \sum_{T\in \mathcal{T}}m_T^q\left(d(\phi(x_T^q), S_f)\right)^2 ,
\end{equation}
where the weights $w_c$ are chosen to be equal to 1, $T$ is a triangle in the set of triangles $\mathcal{T}$ defining the triangle mesh of the ROI, $x_T^q$ is the barycenter of the triangle $T$ and $m_T^q$ is the area of $T$. Thus, the vector $F$ \eqref{eq:VecFGeneral} that encodes the non-linear least squares problem in the 3D setting is
\begin{equation}F[u]= \left[\left\{\sqrt{m_T^q}d(x_T^q+u(x_T^q), S_f)\right\}_{T\in\mathcal{T}}, \lambda M^{-\frac{1}{2}}LU_1,\lambda M^{-\frac{1}{2}}LU_2, \lambda M^{-\frac{1}{2}}LU_3\right]^T.
\end{equation}

The definition of the vector $F$ for the parametric registration is similar to the non-rigid case. The details are presented in Appendix \ref{app-2}.

As in the 2D case, Gauss-Newton requires the solution of a linear system in each iteration of the algorithm. Unlike in the 2D case, it is not feasible though to assemble the system matrix $J_F^TJ_F$ due to memory requirements. $J_F$ does not only have more rows and columns, but also considerably more non-zero entries in each row. Thus, it is crucial not to assemble this product matrix. Instead, we solve the linear system using the LSMR\cite{lsmr2011} algorithm, where it is sufficient to compute and store the matrix $J_F$ and its transposed.

\begin{figure}[b]
\centering
\resizebox{\linewidth}{!}{
\begin{tabular}{cc|c|c|c|c}
\Huge\bf$\lambda=1$&\includegraphics[height = 5.3cm,valign=c]{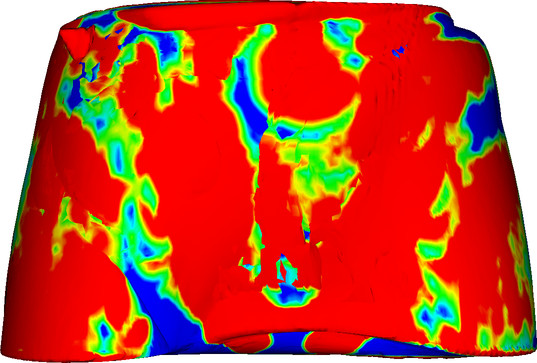}&\includegraphics[height = 5.3cm,valign=c]{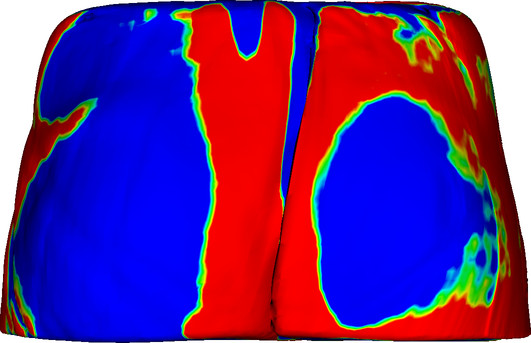}&\includegraphics[height = 5.3cm,valign=c]{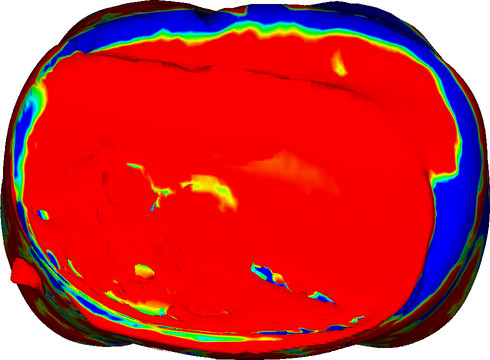}&\includegraphics[height = 5.3cm,valign=c]{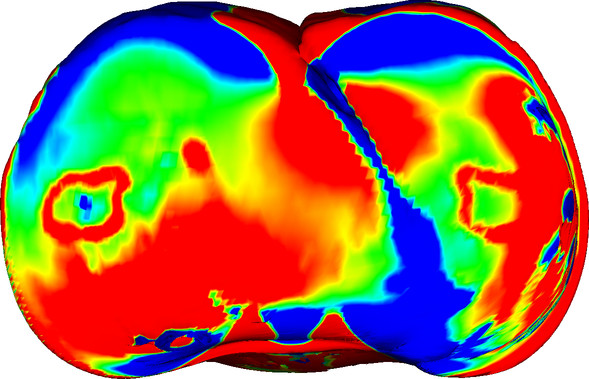}&\includegraphics[height = 5.3cm,valign=c]{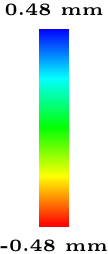}\\
\\ \hline\\
\Huge\bf$\lambda=10^{-1}$&\includegraphics[height = 5.3cm,valign=c]{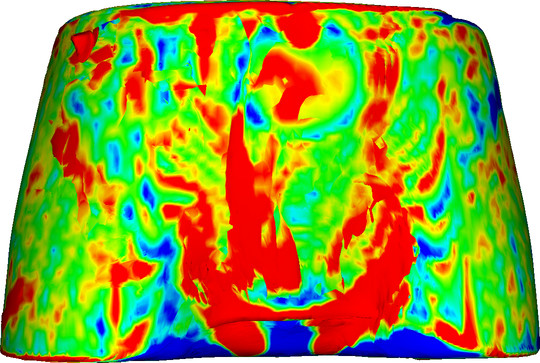}&\includegraphics[height = 5.3cm,valign=c]{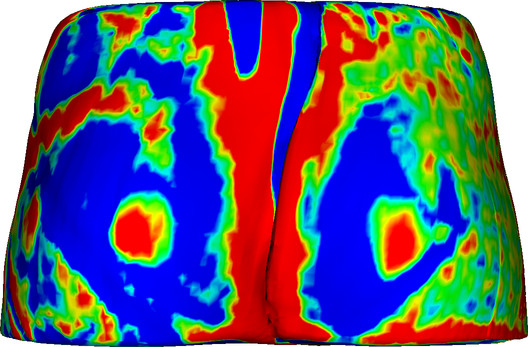}&\includegraphics[height = 5.3cm,valign=c]{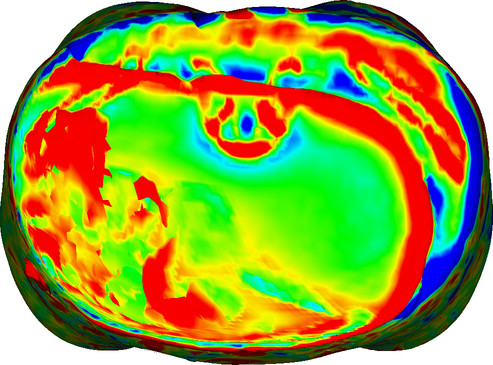}&\includegraphics[height = 5.3cm,valign=c]{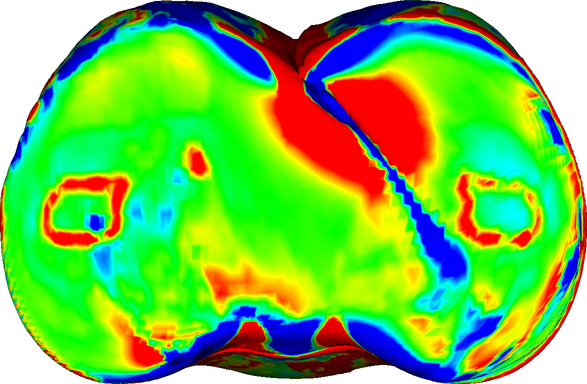}&\includegraphics[height = 5.3cm,valign=c]{colorBar.pdf}\\
\\ \hline\\
\Huge\bf$\lambda=10^{-2}$&\includegraphics[height = 5.3cm,valign=c]{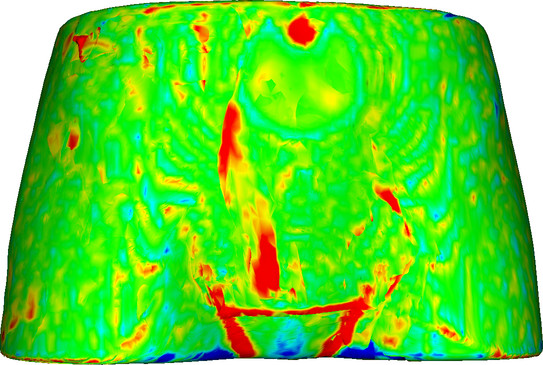}&\includegraphics[height = 5.3cm,valign=c]{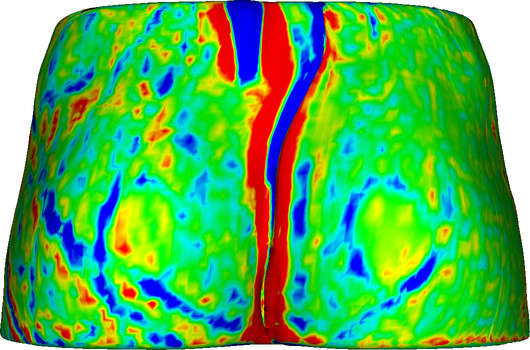}&\includegraphics[height = 5.3cm,valign=c]{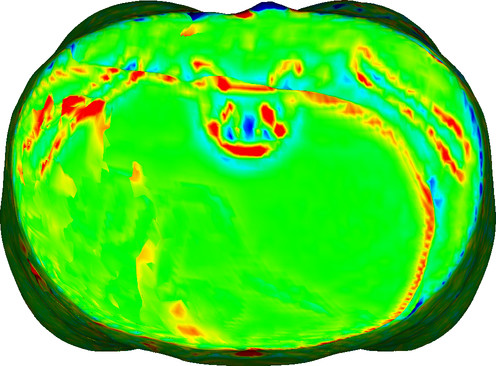}&\includegraphics[height = 5.3cm,valign=c]{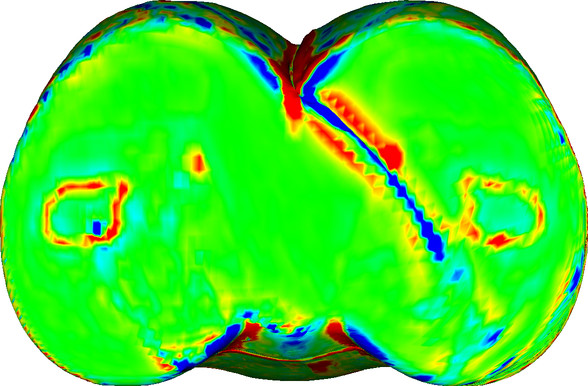}&\includegraphics[height = 5.3cm,valign=c]{colorBar.pdf}\\
\\ \hline\\
\Huge\bf$\lambda=10^{-3}$&\includegraphics[height = 5.3cm,valign=c]{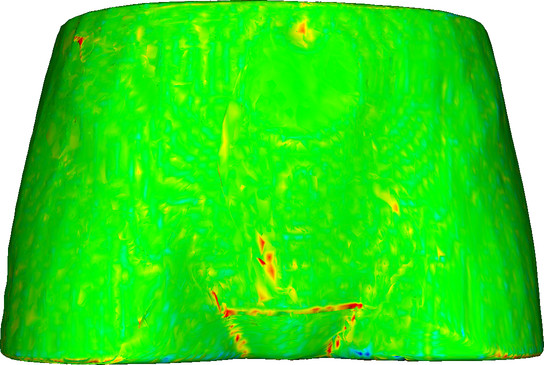}&\includegraphics[height = 5.3cm,valign=c]{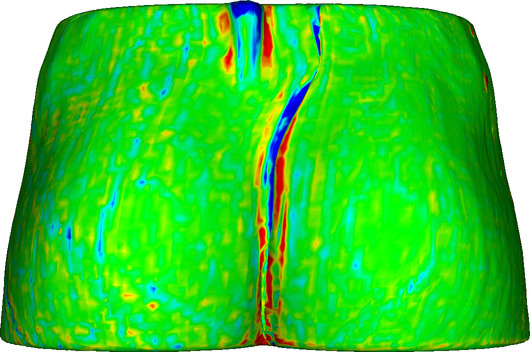}&\includegraphics[height = 5.3cm,valign=c]{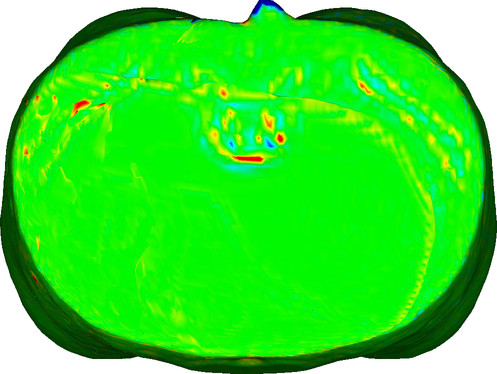}&\includegraphics[height = 5.3cm,valign=c]{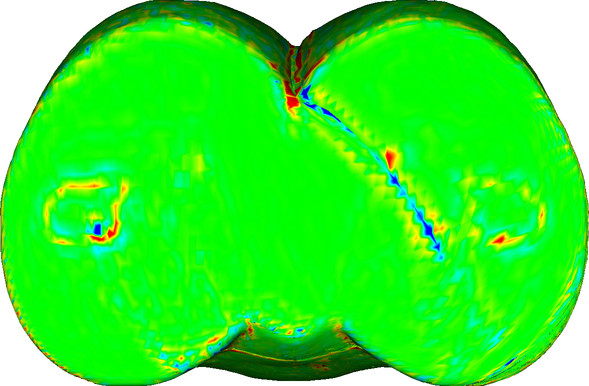}&\includegraphics[height = 5.3cm,valign=c]{colorBar.pdf}\\
\\ \hline\\
\Huge\bf$\lambda=10^{-4}$&\includegraphics[height = 5.3cm,valign=c]{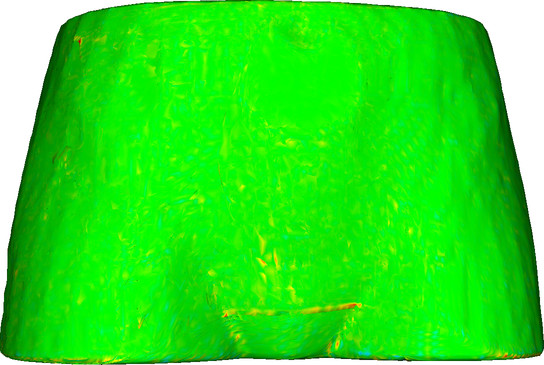}&\includegraphics[height = 5.3cm,valign=c]{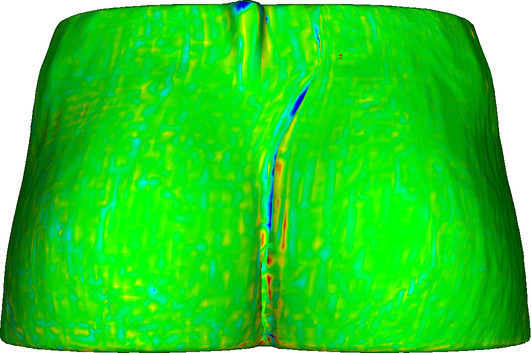}&\includegraphics[height = 5.3cm,valign=c]{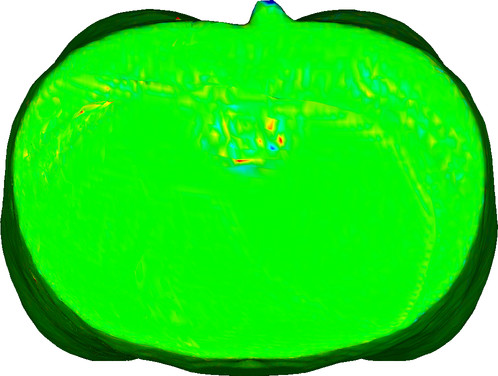}&\includegraphics[height = 5.3cm,valign=c]{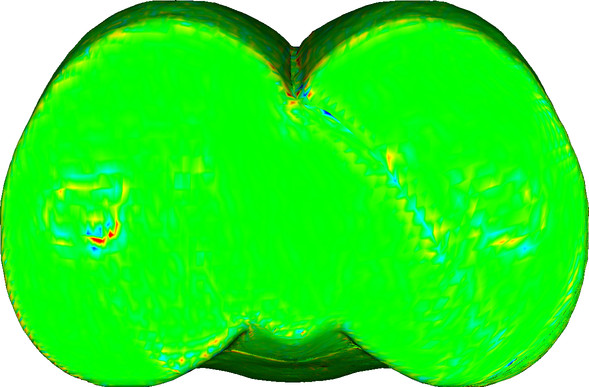}&\includegraphics[height = 5.3cm,valign=c]{colorBar.pdf}\\
\end{tabular}}
\caption{\label{fig:results3Dcolored} The distances $d(\phi(c), S_f)$, for $c\in\mathcal{C}$, are displayed as color coding on the registered template mesh $\phi(\mathcal{C})$ for every value of the parameter $\lambda$ used for the non-rigid registration. From left to right: front, back, top and bottom sides of $\phi(\mathcal{C})$, color bar.}
\end{figure}

Figure \ref{fig:results3Dcolored} illustrates the effect of the minimization strategy proposed in Section~\ref{sec:GenMinimization}. First the parametric registration is performed with the empirically determined parameters $\alpha=1$ and $\mu=1$. Estimating the parameters is rather straightforward since their value mainly depends on the order of magnitude of the initial energy. Then, the non-rigid registration is performed iteratively for decreasing values of the parameter $\lambda$. In this case, the chosen values of the parameters were $\lambda=10^{-i},\;i=\{0,1,2,3,4\}$. This allows to get more accurate registration results as $\lambda$ decreases. For the different settings of $\lambda$, Figure~\ref{fig:results3Dcolored} visualizes the distances $d(\phi(c), S_f)$, for $c\in\mathcal{C}$,
on the registered template mesh, $\phi(\mathcal{C}):=\{\phi(c):c\in\mathcal{C}\}$, for human hips.

\subsection{Evaluation}
\label{sec:Eval}

Reliable valuation of non-rigid registration is difficult, since large databases with reliable ground truth (GT) annotations are not available. Nonetheless, we aim at quantitatively evaluating our approach.

\subsubsection{Aim}

Our evaluation aims at determining the absolute error (accuracy) of the method. To define accuracy, we need to rely on a ROI-based registration problem. This requires automatic segmentation, and the accuracy of the segmentation deeply impacts the overall performance. In addition, we aim at comparing the results to a state-of-the-art method for multi-modal registration, which is considered to be based on mutual information (MI)\cite{oliveira2014}.

\subsection{Metrics}

The alignment accuracy is assessed by the Dice Coefficient\cite{dice} (DC) and the symmetric Hausdorff distance\cite{huttenlocher1993} (HD), which quantify the agreement of two segmentation and the accuracy of the contour alignment, respectively. The larger the DC and the smaller the HD the better the two ROIs correspond.

\subsubsection{Database}

Quantitative measures on ROI alignment are image specific. In order to obtain statistically significant results, a sufficiently large number of images shall be processed. Hence, large databases of images annotated with reliable GT are needed. Therefore, we selected the 2D application of QLF / DP registration, where the ROI is defined as tooth contour.
In total, 150 pairs of QLF and DP of upper and lower incisors and canines have been acquired from 32 subjects. All subjects were exhibiting white spot lesions after orthodontic treatment with a fixed appliance.

\subsubsection{Ground truth}

Manually references are unreliable, since they cannot be reproduced exactly, even with the same rater\cite{SPIE-gold-silver}. However, based on several manual markings, a gold standard can be estimated using the simultaneous truth and performance level estimation (STAPLE) algorithm\cite{staple2004}. The core idea of STAPLE is to iteratively (i) compute the observer-weighted mean of a binary segmentation and (ii) adjust the weights of the observers with respect to the similarity to that mean segmentation. In other words, if a observer has large discrepancies to the estimated GT, the corresponding weights are lowered in the next iteration.

To generate the ground truth with STAPLE, three trained raters (R1, R2, R3) manually marked the tooth contours on both of the image modalities. All the raters were presented the 300 images in random order.

\subsubsection{Assessment of segmentation}

The accuracy of the automatic ROI segmentation is assessed by calculating the DCs and HDs of the automatic segmentation and the ground truth estimated using the STAPLE algorithm. Including the automatic segmentation, in total four observers are available.
The performance of the raters is compared for both modalities, QLF and DP.
The contour extraction from the QLF images is done with the  parameters $\alpha=10^{-3}$, $\mu=10^{-3}$ and $\lambda=10^{-i},\;i=\{4,5,6,7\}$ for the parametric registration step and the non-rigid step, respectively, while for the DP, $\alpha=10^{-3}$, $\mu=10^{-1}$ and $\lambda=10^{-i},\;i=\{4,5,6\}$ were used.

A repeated measures analysis of variance (ANOVA) is applied to assess statistical significances between the automatic segmentation and the human raters, and between the performance of the algorithm in QLF and DP modalities. The significance level is $\alpha = 0.05$.

\subsubsection{Assessment of registration}

The accuracy of the DP/QLF ROI alignment is measured again by DCs and HDs of the photo ground truth deformed using the deformation fields obtained from the proposed registration method and the QLF ground truth.
For the DP/QLF alignment, $\alpha=10^{-3}$, $\mu=10^{-1}$ and $\lambda=10^{-i},\;i=\{4,5,6\}$ were used.

To compare our approach with the state-of-the-art, the same analysis is done using the deformation fields obtained from the Insight Segmentation and Registration Toolkit (ITK)\cite{johnson2015itk} MI registration and those obtained from the MI registration implemented with Elastix\cite{elastix2010}, which is an established registration method.\cite{monti2017,broggi2017,malinsky2013} In both cases, the registration is performed on the same grayscale version of the DP/QLF pairs, which are used for the proposed automatic segmentation. The parameters used for the registrations are specified in Table \ref{tab:paramMI}. The listed parameters are the default values suggested by the original authors, except for the ``Final Grid spacing" parameter for Elastix. The latter has been increased from its default value of 16 pixels to 48 pixels, as using the default value was resulting in obviously unrealistic transformations. In the ITK implementation, the MI is optimized using the Limited-memory Broyden Fletcher Goldfarb Shannon\cite{lbfgsb} (L-BFGS-B) algorithm.

Based on DC and HD, a one-way repeated measures ANOVA is used to compare the proposed method with the mutual information-based ITK implementation that is considered as gold standard.

\begin{table}[t]
\caption{\label{tab:paramMI} Parameters used for ITK and Elastix. Here, MI indicates mutual information and L-BFGS-B refers to Limited-memory Broyden Fletcher Goldfarb Shannon algorithm.}
\vspace{0.3cm}
\centering
\begin{tabular}{l|l|l}\hline
{\bf Parameter} &\bf{ITK}&\bf{Elastix}\\
\hline
Metric & Mattes MI & Advanced Mattes MI\\
Number of histogram bin & 50 & 32 \\
Transformation & Third-order B-spline & Affine + Third-order B-spline \\
Final Grid spacing & \# pixels in the input image & 48 pixels \\
Optimization algorithm & L-BFGS-B & Adaptive stochastic gradient\\
& & descent\\
Maximum number of iterations & 1000 & 200 (Affine) - 500 (B-spline)\\
Number of multiresolution levels & - & 4 \\
\hline
\end{tabular}
\end{table}

\section{Results}
\label{sec:Res}

In this section, we present the results of qualitative registration in 2D and 3D, as well as the accuracy determined in the  2D evaluation for segmentation and registration.

\subsection{2D application}
\begin{figure}[htb]
\centering
\resizebox{\linewidth}{!}{\begin{tikzpicture}
 \node at (-3,0){\includegraphics[height=5cm]{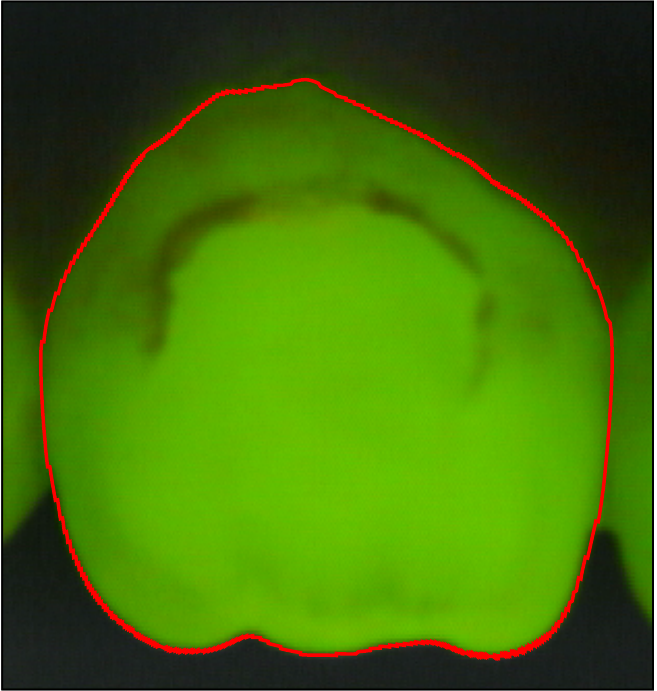}};
 \node[fill=white,circle,inner sep = 0.6ex] at (-5,-2.1) { \bf \textcolor{black}{a}};
 \node at (2,0){\includegraphics[height=5cm]{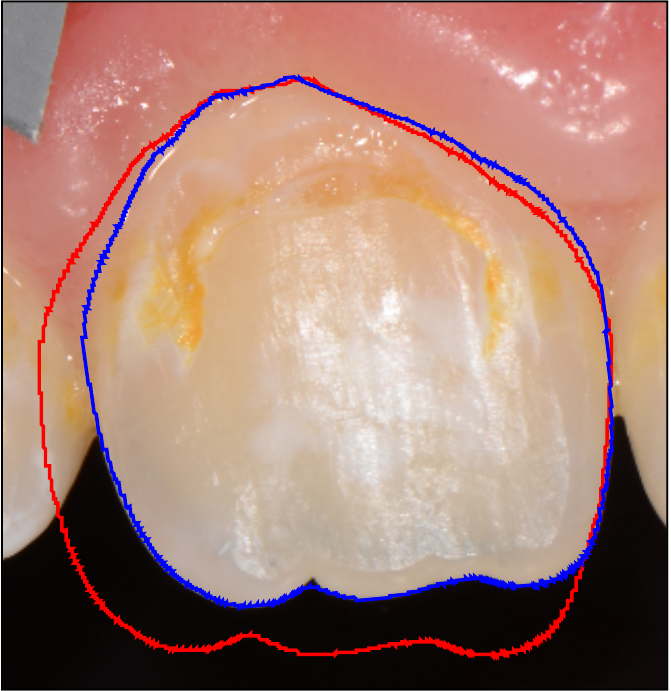}};
 \node[fill=white,circle,inner sep = 0.6ex] at (0,-2.1) { \bf \textcolor{black}{b}};
 \node at (7,0){\includegraphics[height=5cm]{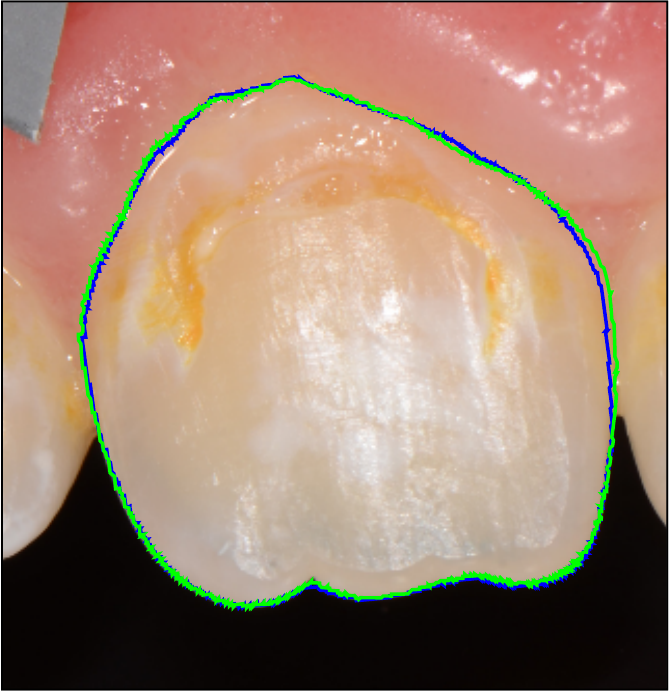}};
 \node[fill=white,circle,inner sep = 0.6ex] at (5,-2.1) { \bf \textcolor{black}{c}};
 \node at (12,0){\includegraphics[height=5cm]{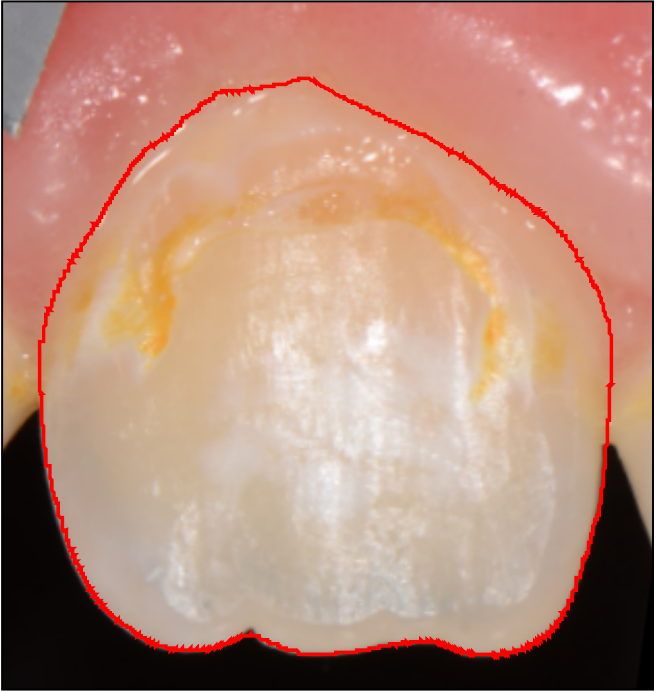}};
 \node[fill=white,circle,inner sep = 0.6ex] at (10,-2.1) { \bf \textcolor{black}{d}};
 \node at (17,0){\includegraphics[height=5cm]{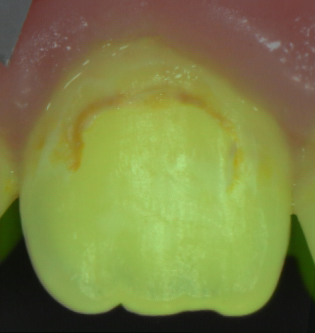}};
 \node[fill=white,circle,inner sep = 0.6ex] at (15,-2.1) { \bf \textcolor{black}{e}};
\end{tikzpicture}}
\caption{\label{fig:figRes} Results in 2D: The QLF image with the the superimposed curve (red) representing the automatic segmentation (a); the DP with superimposed curves (b); QLF-based tooth contour before (red) and after parametric registration (blue); and the curves after parametric and non-rigid registration in blue and green, respectively (c). Panel (d) shows the DP after non-rigid deformation and the corresponding QLF curve (red). Finally, an image created by blending the QLF and the registered DP is displayed in (e).}
\end{figure}
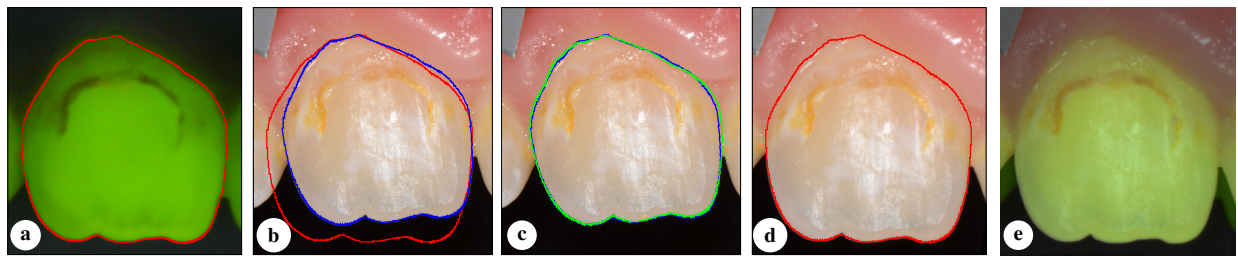

Figure \ref{fig:figRes} shows qualitative results obtained in the exemplary 2D application. After non-rigid deformation, the DP matches the QLF, as the QLF-based contour (red) matches the tooth in the DP, as depicted in Panel (d).

\subsection{3D application}
\label{subsec:3DMethod}

\begin{figure}[t]
\centering
\resizebox{0.8\linewidth}{!}{
\begin{tabular}{l | l |  l}
\includegraphics[height = 5.7cm]{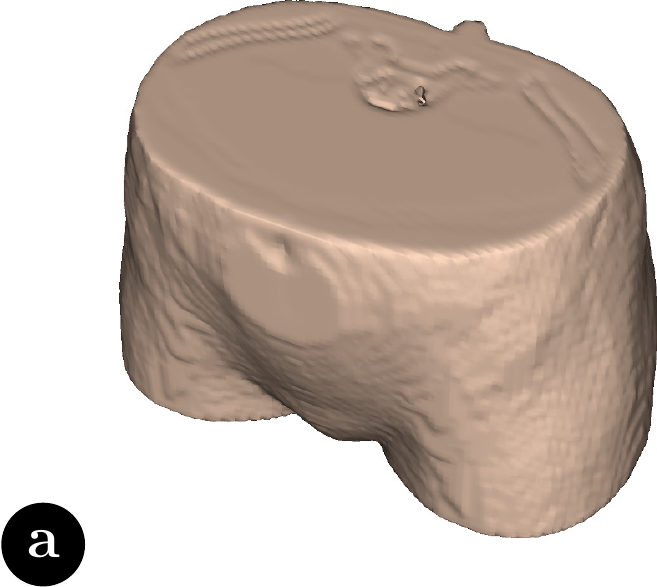}&\includegraphics[height = 5.6cm]{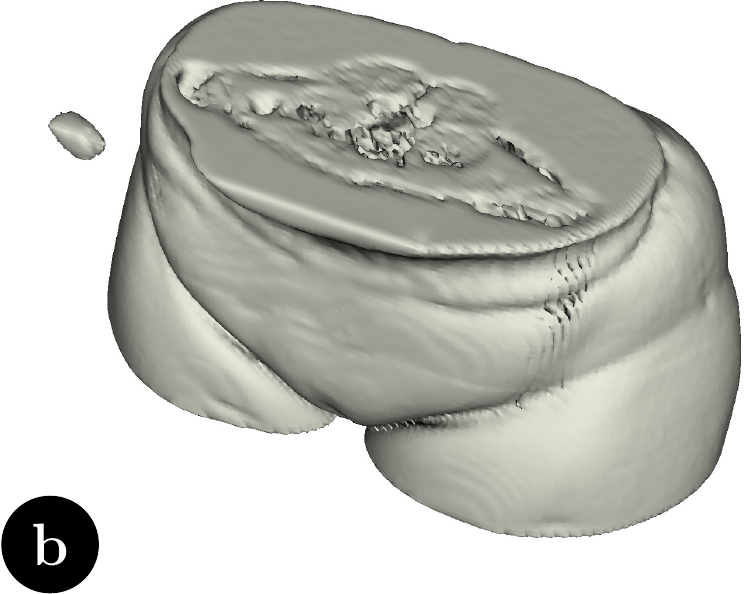}&\includegraphics[height = 5.7cm]{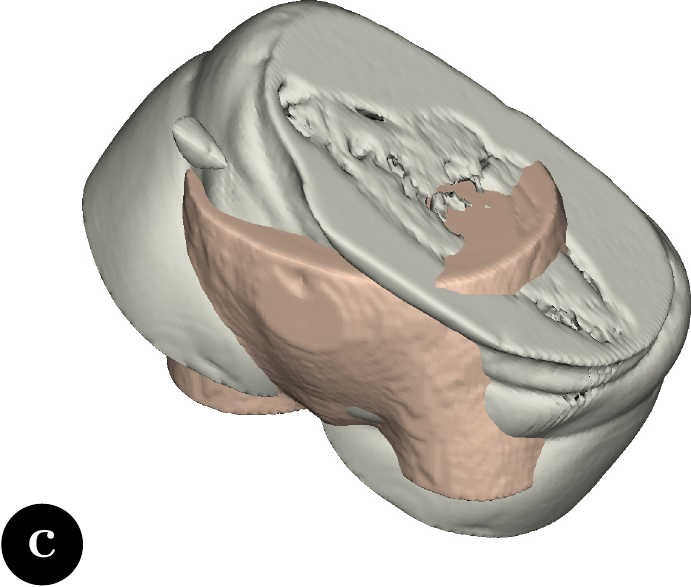}\\
\\ \hline\\
\includegraphics[height = 5.7cm]{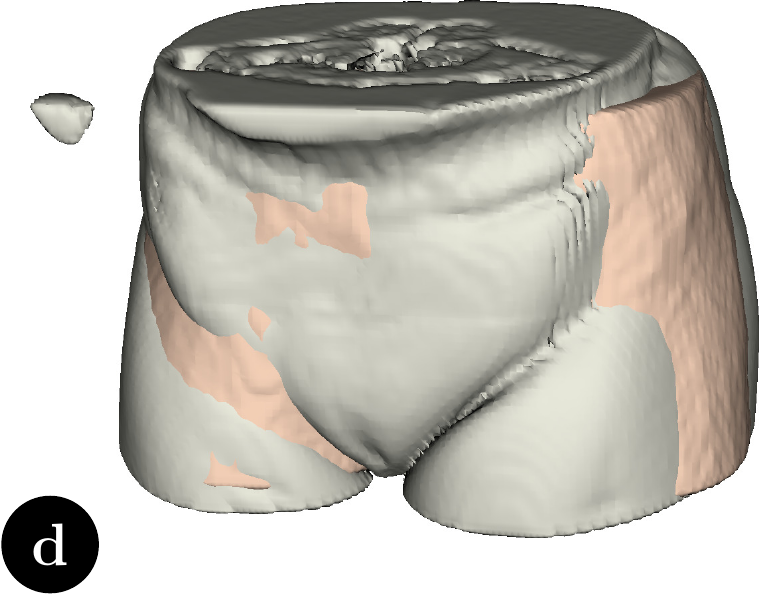}&\includegraphics[height = 5.7cm]{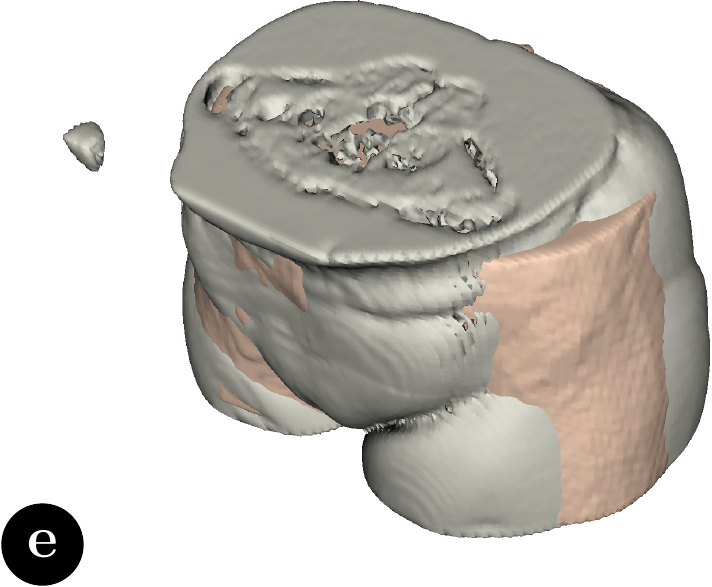}&\includegraphics[height = 5.7cm]{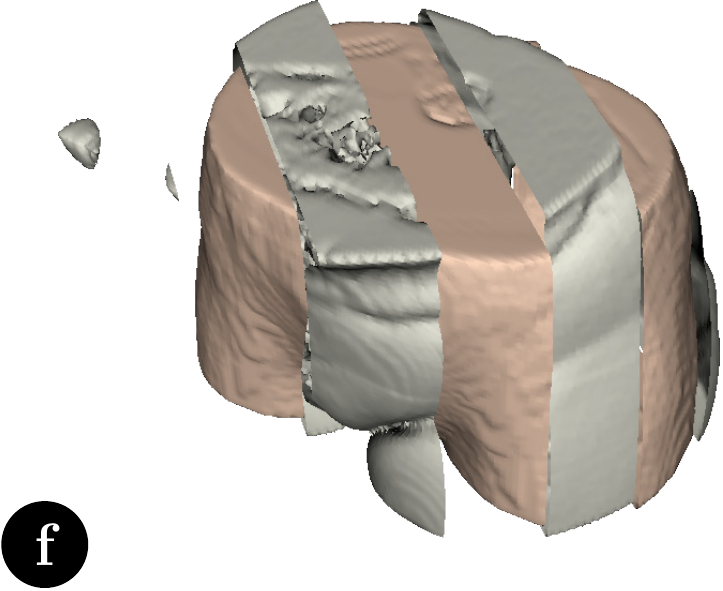}\\
\\ \hline\\
\includegraphics[height = 5.7cm]{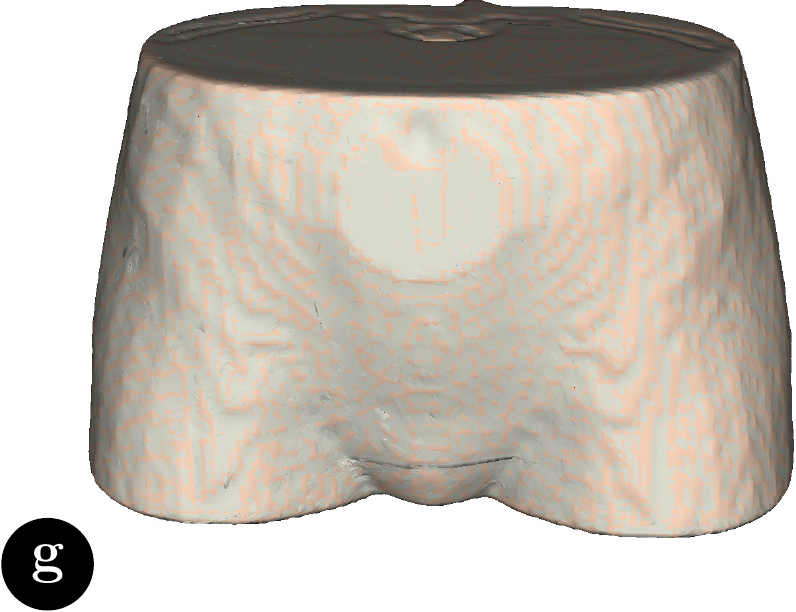}&\includegraphics[height = 5.7cm]{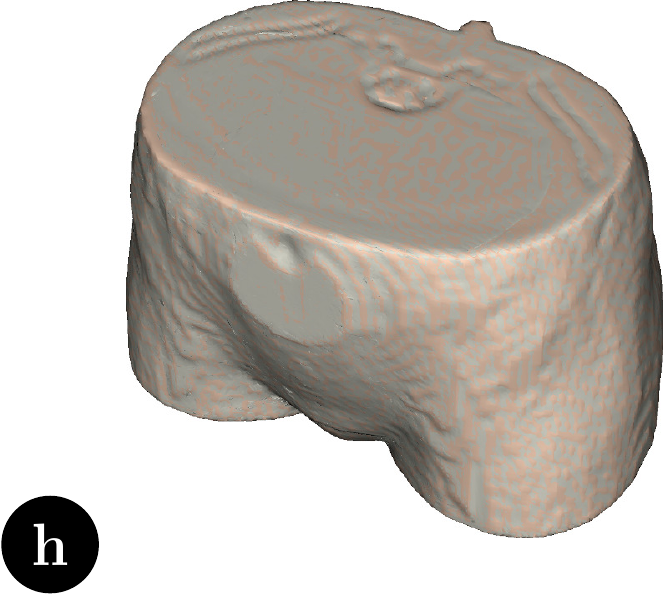}&\includegraphics[height = 5.7cm]{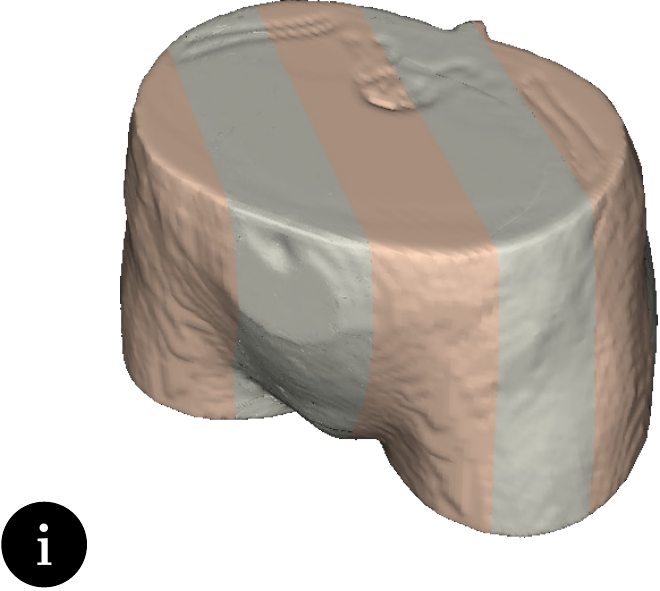}\\
\end{tabular}}
\caption{\label{fig:results3D} Results in 3D.
The first row shows the starting setting: $S_f$ (a), template mesh $g$ (b), and initial position of the input images (c). Note that (a) and (b) use different manually chosen view angles to simplify the comparison, (c) shows the true initial mismatch of the data sets. Middle and lower row depict the results after parametric registration and non-rigid deformation, respectively. Panels (f) and (j) visualize the according slices of $S_f$ and of the registered template mesh after parametric and non-rigid registration respectively.}
\end{figure}

The M2R registration method has been applied in 3D, too.
For the use case from the RASimAs project, the resolution of the corresponding 3D grid is $129\times 129\times129$, with pixel dimensions (3.73, 3.73, 3.73) mm. Figure \ref{fig:results3D} depicts the results and suggests an appropriate surface-to-voxel alignment. The Dice coefficient of the two meshes after the alignment is 0.9959. As measure of the distance between the contours, the Hausdorff distance is calculated. The Hausdorff distance of the reference mesh from the template mesh is 90.60 mm (24.28 pixel units) before performing the registration and 15.73 mm (4.22 pixel units) after. The 99th percentile of the Hausdorff distance of the reference from the template is 78.67 mm (21.14 pixel units) before and 2.12 mm (0.57 pixel units) after the registration. Note that the 99th percentile after the registration is much lower than the  Hausdorff distance after the registration. This indicates the presence of just very few outlier points (Fig. \ref{fig:results3Dcolored}, blue/red dots in the last row) that make the Hausdorff distance appear relatively large even though the alignment is very accurate. Furthermore, note that we are not considering the symmetric Hausdorff distance here since one surface only contains part of the other surface. The proposed method, in fact, has the advantage that it can cope with non-symmetric relations between the two input meshes.

\subsection{Quantitative evaluation of segmentation performance}

Based on the 150 QLF images, the mean of the DC between the segments obtained from the automatic procedure and the STAPLE-based GT that is calculated from the three human observers is 0.978 (range 0.9420-0.9940). The mean HD is 0.032 (range 0.0102-0.1060).

Based on the 150 DP images, these values turn to 0.981 (range 0.9440-0.9940) and 0.030 (range 0.0102-0.0792) for DC and HD, respectively (Table \ref{tab:compGTQLFDP}).

For both QLF and DP, a pairwise comparison of the means of the DC as well as of the HD using a one-way repeated measure ANOVA was performed between the automatic segmentation and each rater, determining a statistical significant difference between these means (Table \ref{tab:ANOVA}).

The one-way repeated measures ANOVA revealed a statistical significant difference between the means of the DC ($F(1,149)=8.68$, $p=0.0037$) between DP and QLF modalities, but no significant difference between the means of the HD ($F(1,149)=1.86$, $p=0.1742$).

\begin{table}[t]
\caption{\label{tab:compGTQLFDP} Mean and standard deviation (SD) of Dice coefficient (DC) and symmetric Hausdorff distance (HD) of the ground truth and manual markings and of ground truth and the proposed automatic segmentation (M2R) for QLF and DP.}
\vspace{0.3cm}
\centering
\begin{tabular}{c|cc|cc|cc|cc}\hline
{\bf Modality} &\multicolumn{4}{c|}{QLF}&\multicolumn{4}{c}{DP}\\
\hline
{\bf Metrics} & \multicolumn{2}{c}{DC} & \multicolumn{2}{c|}{HD}& \multicolumn{2}{c}{DC} & \multicolumn{2}{c}{HD}\\
& Mean & SD & Mean & SD & Mean & SD & Mean & SD \\
\hline
{\bf Method}& &     &        &        &        &        &        &       \\
Rater R1  & 0.990        & 0.0098 & 0.015 & 0.0117 & 0.984        & 0.0076 & 0.021 & 0.0087\\
Rater R2  & 0.980        & 0.0068 & 0.023 & 0.0067 & 0.991        & 0.0066 & 0.015 & 0.0094\\
Rater R3  & {\bf 0.992}  & 0.0082 & 0.012 & 0.0116 & {\bf  0.996} & 0.0049 & 0.008 & 0.0079\\
M2R & 0.978        & 0.0101 & 0.032 & 0.0185 & 0.981        & 0.0083 & 0.030 & 0.0129\\
\hline
\end{tabular}
\end{table}

\subsection{Quantitative evaluation of registration performance}

The descriptive statistics (mean and standard deviation) of DCs and HDs measuring the accuracy of the alignment for the ITK-MI and Elastix state-of-the-art methods and our mesh-to-raster (M2R) approach are reported in Table \ref{tab:compGTdef}.
Based on the 150 QLF/DP image pairs, the means of the DC are 0.940 (range 0.7600-0.9900), 0.959 (range 0.874-0.991) and 0.971 (range 0.8910-0.9930) for ITK-MI, Elastix and M2R, respectively. The means of the HD are 0.072 (range 0.0181-0.2630), 0.055 (range 0.014-0.153)
and 0.041 (range 0.0102-0.1050), respectively.

\begin{figure}[t]
\centering
\includegraphics[height=5cm]{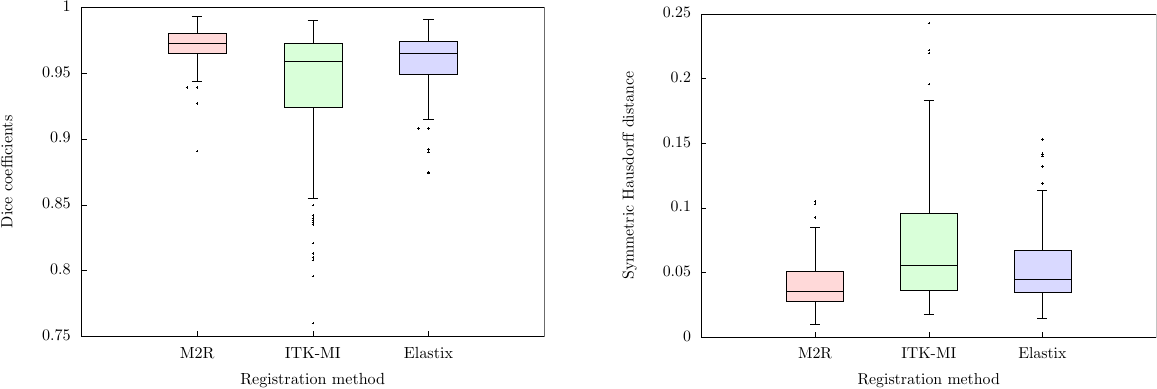}
\caption{\label{fig:BoxplotQLFVI} Boxplots of the Dice coefficients and the symmetric Hausdorff distance of the QLF ground truth and the DP ground truth deformed using the proposed algorithm (M2R), the ITK mutual information (ITK-MI) and the registration implemented with Elastix.}
\end{figure}

The one-way repeated measures ANOVA revealed a statistical significant difference between the means of the DC ($F(1,149)=71.96$, $p<0.001$) as well as the means of the HD ($F(1,149)=61.99$, $p<0.001$) for the registration with ITK and the proposed method. Similar results were given by ANOVA for the means of the DC ($F(1,149)= 62.96$, $p<0.001$) and for the means of the HD ($F(1,149)=38.14$, $p<0.001$) for M2R and Elastix. The statistical significance is emphasized by the box plots in Figure \ref{fig:BoxplotQLFVI}, which are visualizing the data from Table \ref{tab:compGTdef}.
Both the DCs and the HDs exhibit more variability in the case of ITK-MI and Elastix rather than for our method.

\begin{table}[t]
\caption{\label{tab:compGTdef} Mean and standard deviation (SD) of Dice coefficient (DC) and symmetric Hausdorff distance (HD) of the deformed DP ground truth and the QLF ground truth for the proposed mesh-to-raster (M2R), the ITK-MI and Elastix reference methods.}
\vspace{0.3cm}
\centering
\begin{tabular}{c|cc|cc}
\hline
{\bf Metrics} & \multicolumn{2}{c}{DC} & \multicolumn{2}{c}{HD}\\
        & Mean & SD & Mean & SD \\
\hline
{\bf Method} &  &        &        &       \\
\textbf{M2R }   & {\bf 0.971} & 0.0129 & {\bf 0.041} & 0.0180\\
ITK-MI & 0.940       & 0.0471 & 0.072       & 0.0494\\
Elastix & 0.959 & 0.0219 & 0.055 & 0.0279\\
\hline
\end{tabular}
\end{table}

\section{Discussion}
The alignment of ROIs in multi-modal medical images is important for many applications. Here, a mesh-to-raster (M2R)-based method is described, which aligns the contour of the ROI of the reference image to the target image with a deformation field that is determined on the whole image domain.
Although this paper provides examples in 2D and 3D only, the method is applicable in general in any dimension.

The quantitative evaluation is based on multi-modal 2D data from dentistry. Our methods was applied to QLF and DP images, including an evaluation of the automatic segmentation that is needed to determine the tooth region, which acts as ROI.

With respect to the evaluation of the ROI segmentation, our M2R approach is not as accurate as a human observer (a significant difference among the means of each human rater and the automatic segmentation was determined by ANOVA). In absolute numbers for QLF, the distance in terms of mean DC between M2R ($ {\rm DC} = 0.978$) and the best-performing human observer R3 ($ {\rm DC} = 0.992$) is only about $1.5 \%$, while the distance between the M2R and the worst-performing human observer R2 is negligible (about $0.2\%$). Similar results hold for the DP.
Also, the mean of the HD for the automatic segmentation is slightly bigger than those of the manual markings.

The observed deviation from the GT in a QLF is due to the extended and smooth transition zone between tooth and background, leading to an imprecise classification of the image into tooth and non-tooth regions and thus to a less accurate contour extraction than in cases where the images exhibit clear distinction between tooth and background. Similarly for the DP images, small inaccuracies in tooth and non-tooth regions classification are caused mostly by the low contrast between the colors of tooth and gum or adjacent teeth.
However, the segmentation step may be replaced easily by another algorithm.

Contour-based registration methods have been used in several works for ROI extraction and alignment. Chen et al.\cite{chen2005} used an automatic method to extract and align teeth contours to register dental radiographs. A deformable image registration method that used ROI's contour propagation was proposed by Wu et al.\cite{wu2015} in radiotherapy. Here, the proposed method is used for both ROI extraction and matching. The method is correctly aligning the contour of the ROI in the target image to the boundary of the ROI in the reference image. However, if the boundary of ROI in the reference image is not accurately delineated, it might lead to inaccuracies in the registration step, as already pointed out in the case of QLF and DP segmentation.

\begin{figure}
\centering
\resizebox{\linewidth}{!}{\begin{tabular}{cccccccc}
\toprule
\makecell{\Large{\bf{QLF with}}\\ \Large{\bf{marked GT}}} & \makecell{\Large{\bf{DP with}}\\ \Large{\bf{marked GT}}} & \makecell{\Large{\bf{DP deformed}}\\ \Large{\bf{using M2R}}} & \makecell{\Large{\bf{DP deformed}}\\ \Large{\bf{with ITK-MI}}}& \makecell{\Large{\bf{DP deformed}}\\ \Large{\bf{with Elastix}}} & \makecell{\Large{\bf{Quality measure}}\\ \Large{\bf{for M2R}}} & \makecell{\Large{\bf{Quality measure}}\\ \Large{\bf{for ITK-MI}}}& \makecell{\Large{\bf{Quality measure}}\\ \Large{\bf{for Elastix}}}\\
\midrule
\includegraphics[height=4cm,valign=c]{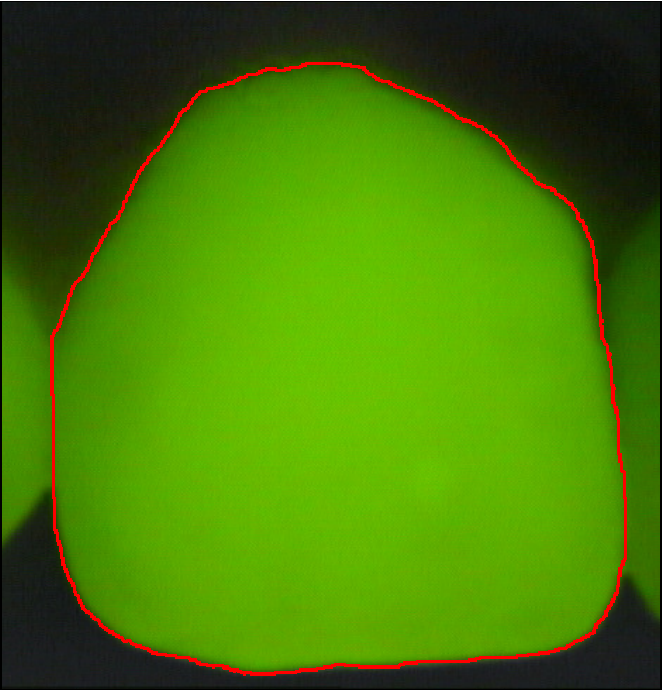}&\includegraphics[height=4cm,valign=c]{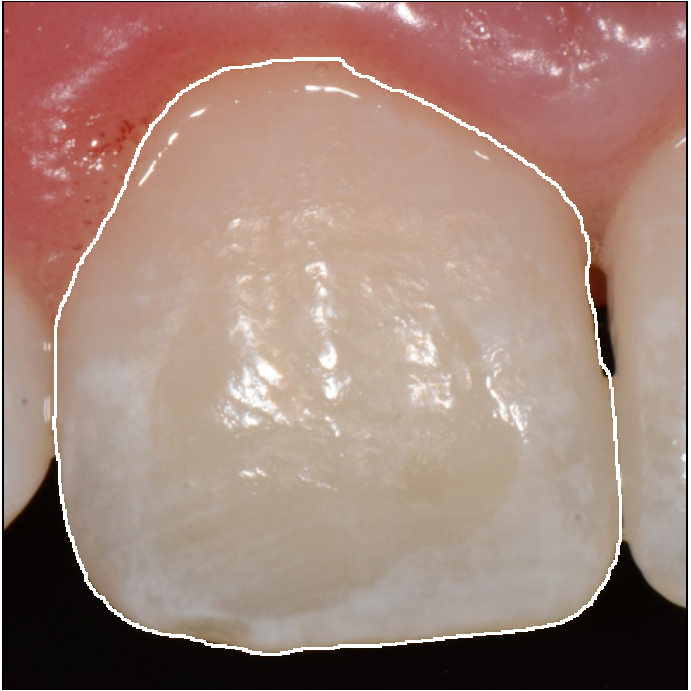}&\includegraphics[height=4cm,valign=c]{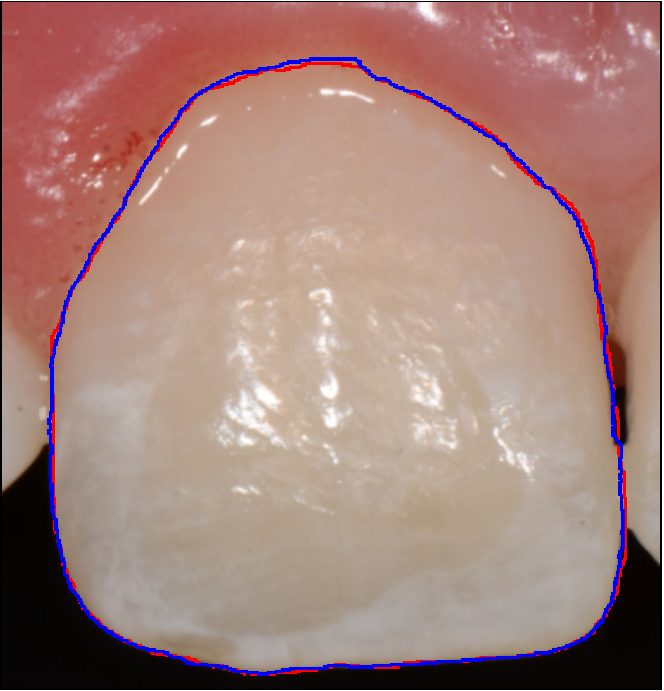}&\includegraphics[height=4cm,valign=c]{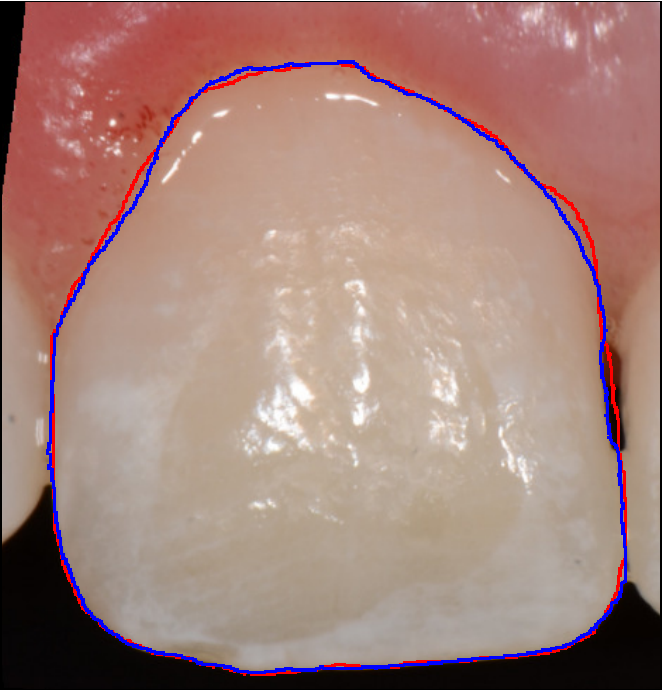}&\includegraphics[height=4cm,valign=c]{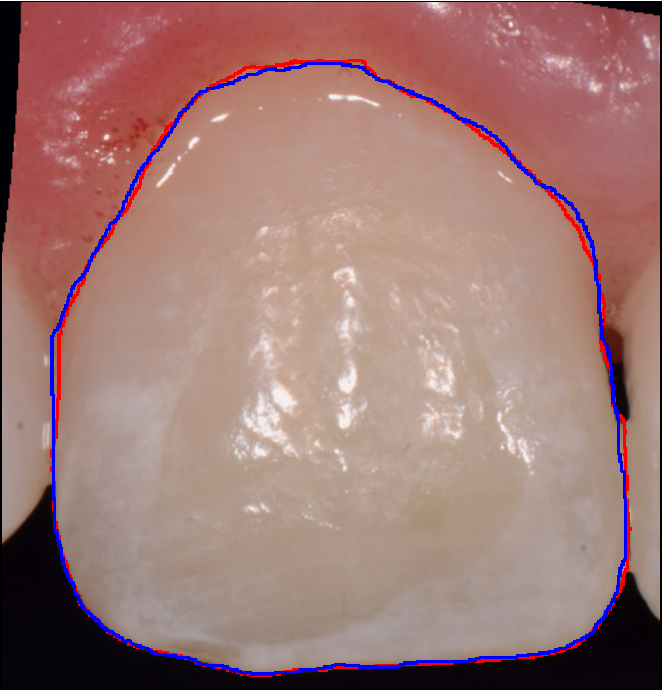} & \makecell{\Large{Best DC (0.9930)} \\ \Large{Best HD (0.0102)}} & \makecell{\Large{Best DC (0.9900)}\\ \Large{      \hspace{22pt}   HD (0.0201)}}& \makecell{\Large{Best DC (0.9910)}\\ \Large{Best HD (0.0145)}} \\
\midrule
\includegraphics[height=4.7cm,valign=c]{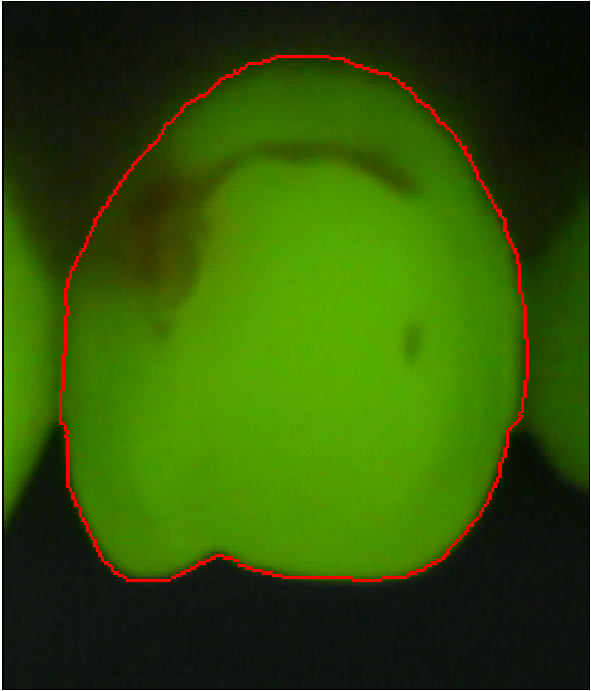}&\includegraphics[height=4.7cm,valign=c]{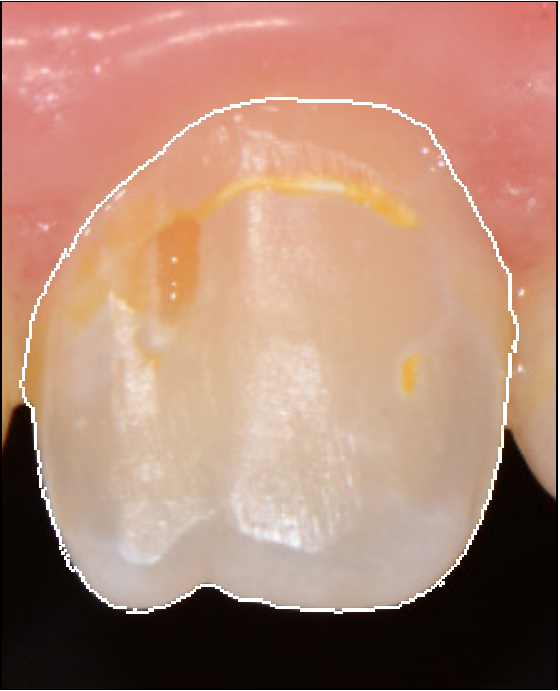}&\includegraphics[height=4.7cm,valign=c]{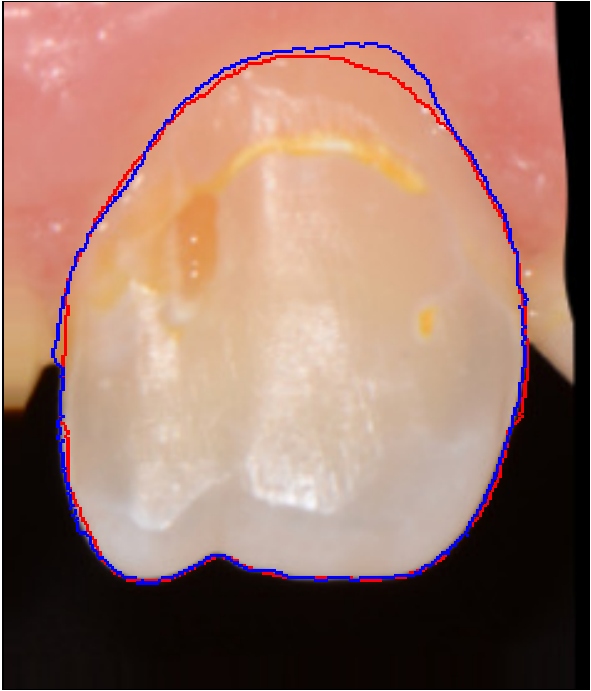}&\includegraphics[height=4.7cm,valign=c]{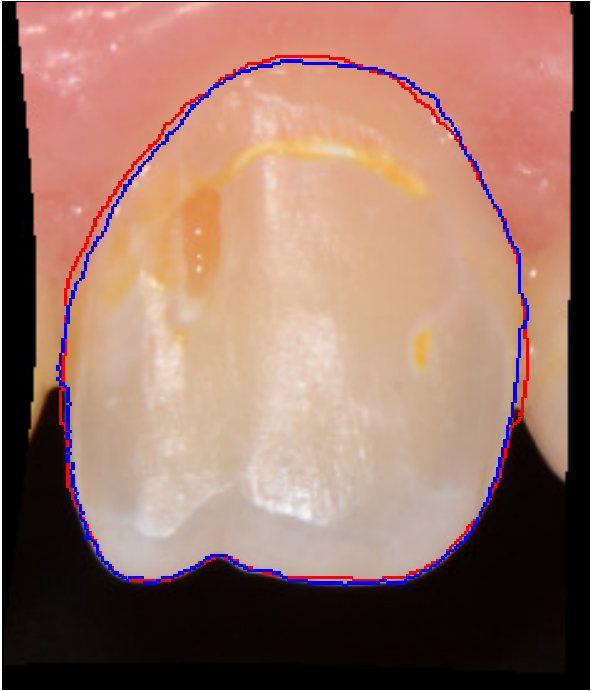}&\includegraphics[height=4.7cm,valign=c]{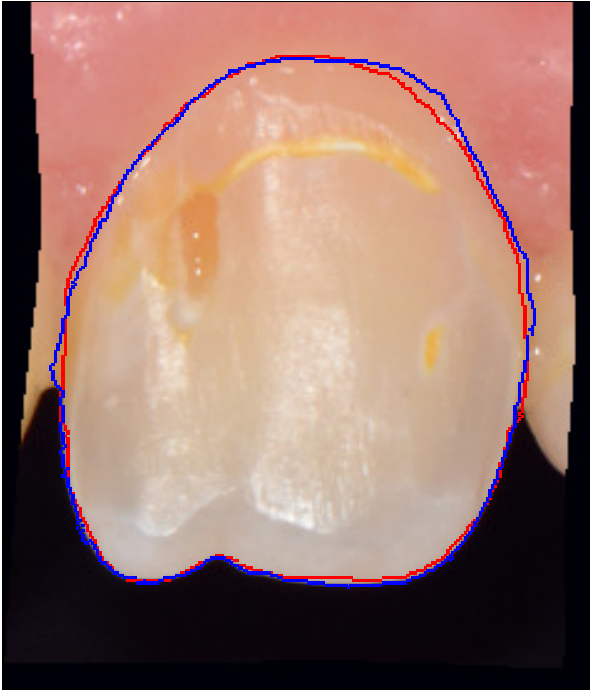}& \makecell{\Large{DC (0.9750)}\\ \Large{HD (0.0397)}}&\makecell{\Large{ \hspace{26pt} DC (0.9820)}\\ \Large{  Best HD (0.0181)}}&\makecell{\Large{DC (0.9830)}\\ \Large{HD (0.0255)}}\\
\midrule
\includegraphics[height=3.9cm,valign=c]{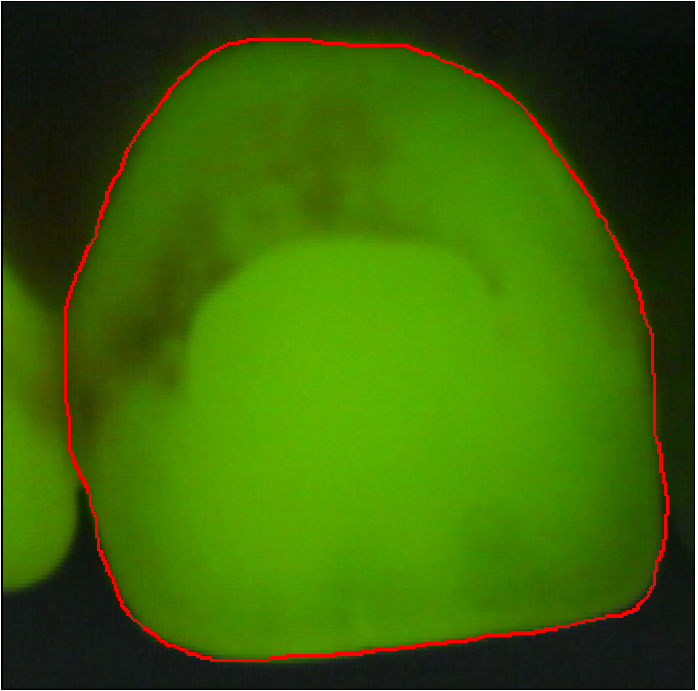}&\includegraphics[height=3.9cm,valign=c]{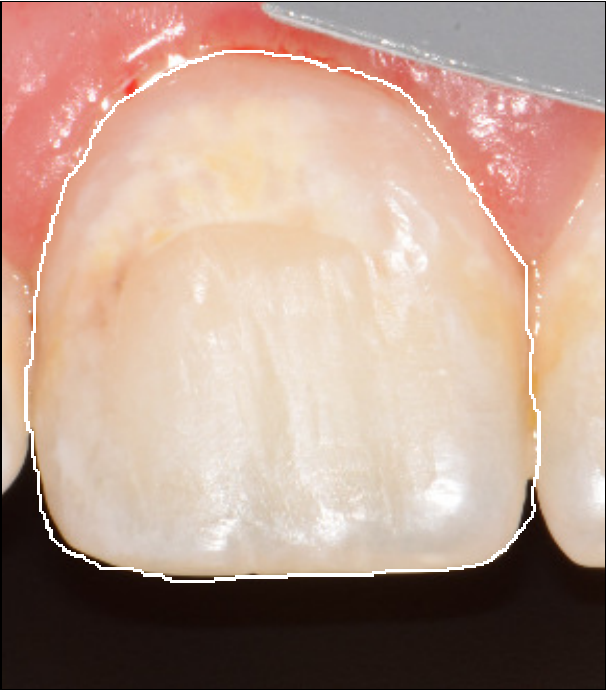}&\includegraphics[height=3.9cm,valign=c]{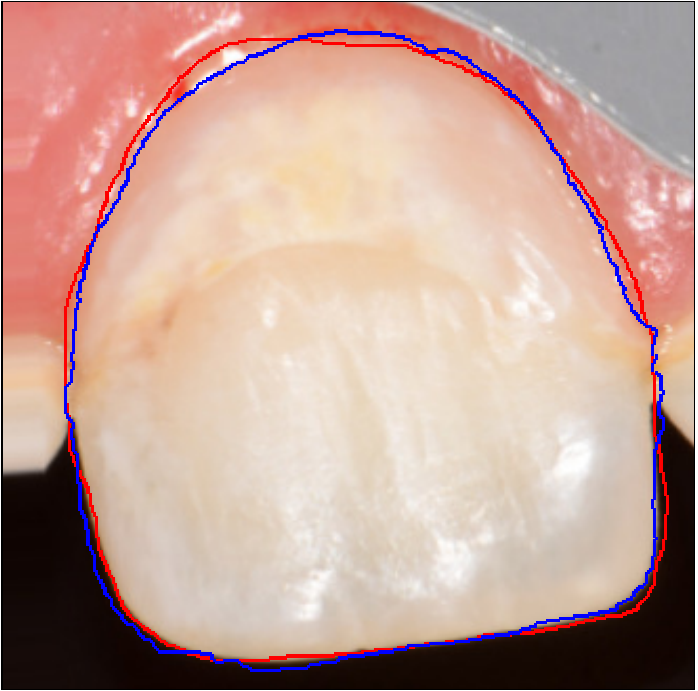}&\includegraphics[height=3.9cm,valign=c]{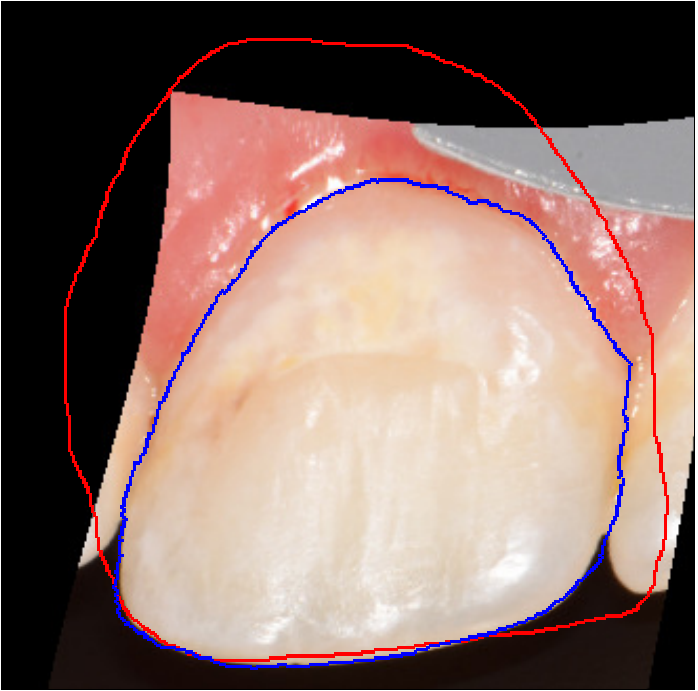}&\includegraphics[height=3.9cm,valign=c]{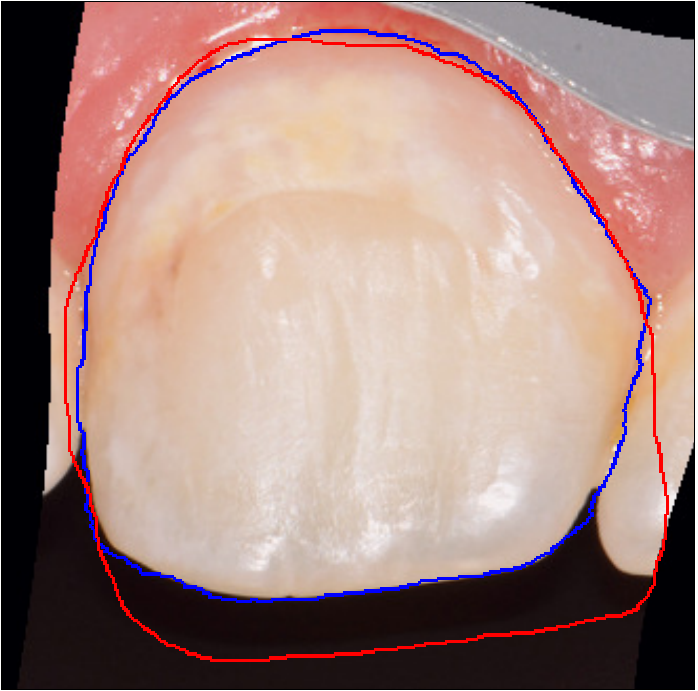}&\makecell{\Large{DC (0.9760)} \\ \Large{HD (0.0322)}} & \makecell{\Large{Worst DC (0.7600)}\\ \Large{Worst HD (0.2630)}}& \makecell{\Large{DC (0.9080)}\\ \Large{HD (0.1420)}}\\
\midrule
\includegraphics[height=4.8cm,valign=c]{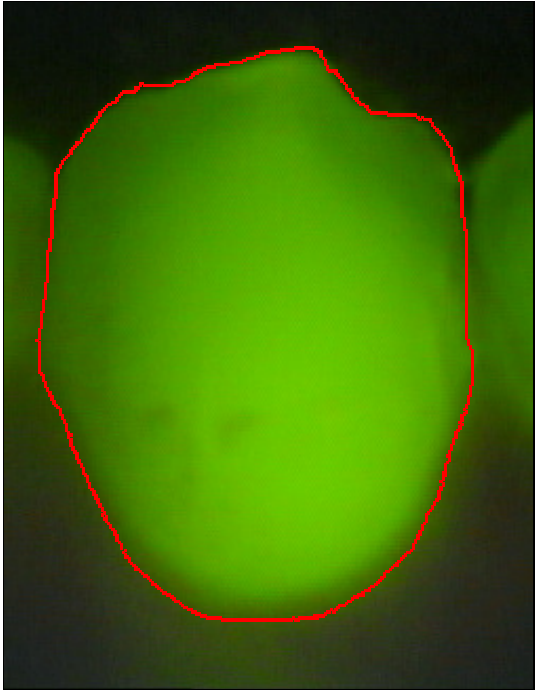}&\includegraphics[height=4.8cm,valign=c]{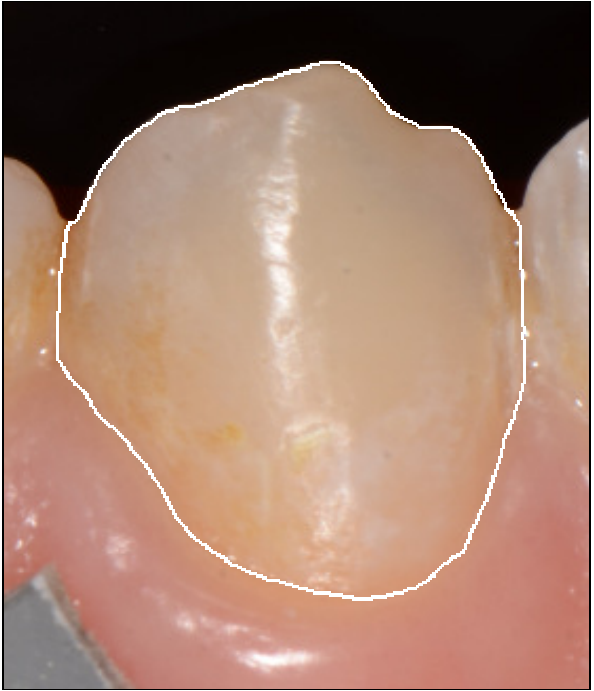}&\includegraphics[height=4.8cm,valign=c]{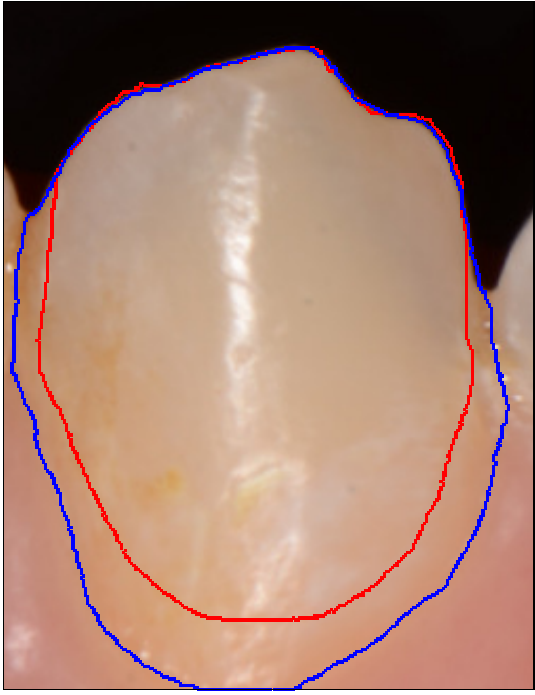}&\includegraphics[height=4.8cm,valign=c]{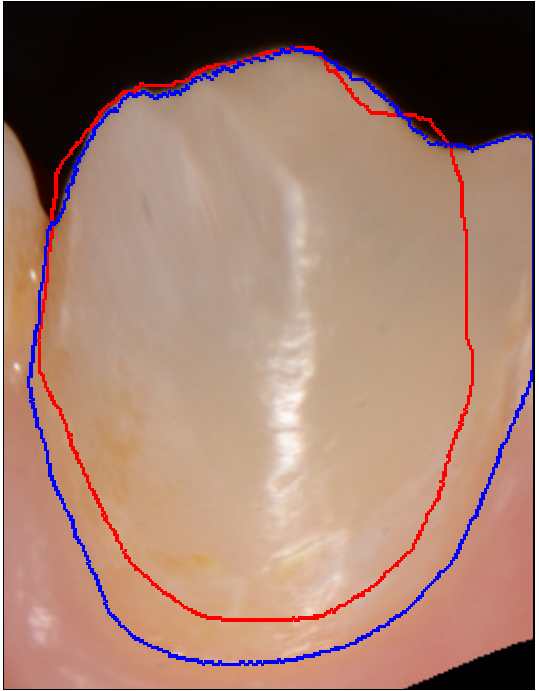}&\includegraphics[height=4.8cm,valign=c]{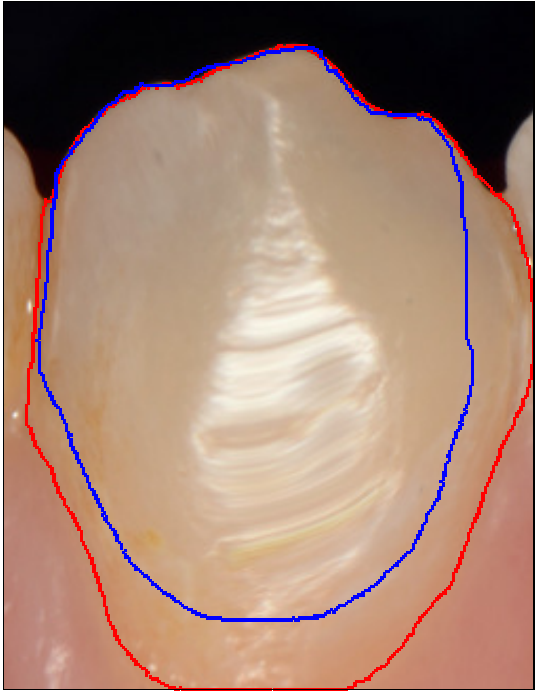}& \makecell{\Large{Worst DC (0.8910)} \\ \Large{Worst HD (0.1050)}} &\makecell{\Large{DC (0.8810)} \\ \Large{HD (0.1190)}}& \makecell{\Large{Worst DC (0.8740)} \\ \Large{      \hspace{25pt}     HD (0.1140)}} \\
\midrule
\includegraphics[height=4.8cm,valign=c]{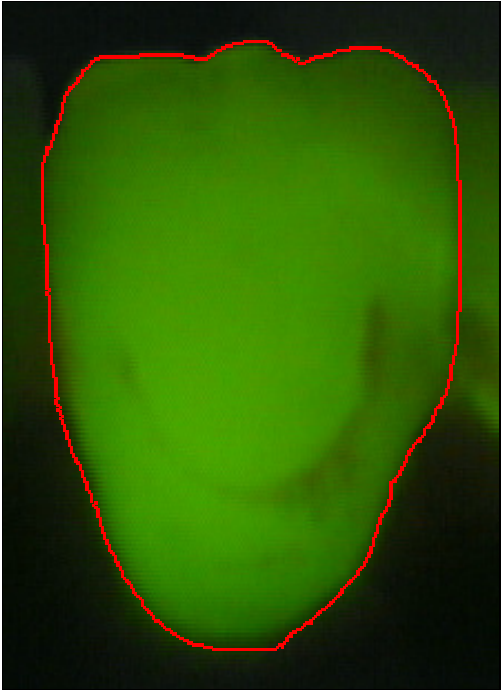}&\includegraphics[height=4.8cm,valign=c]{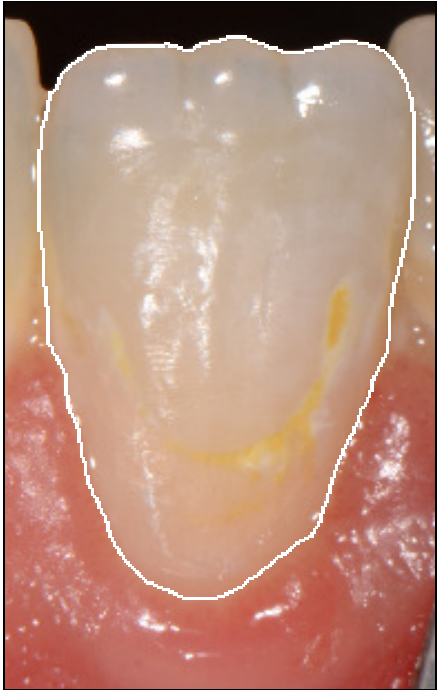}&\includegraphics[height=4.8cm,valign=c]{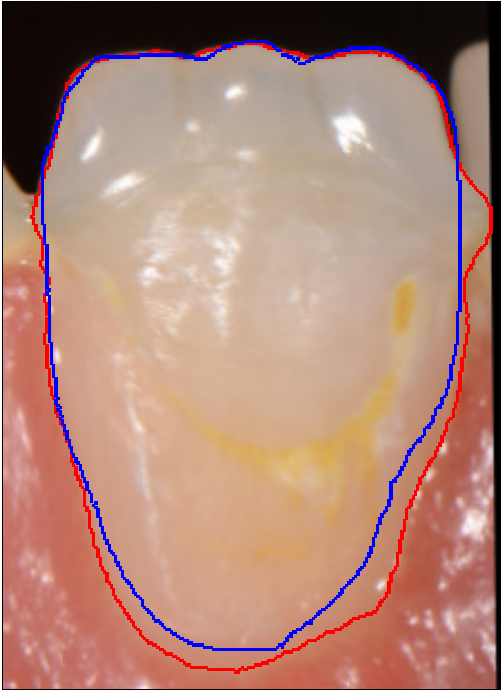}&\includegraphics[height=4.8cm,valign=c]{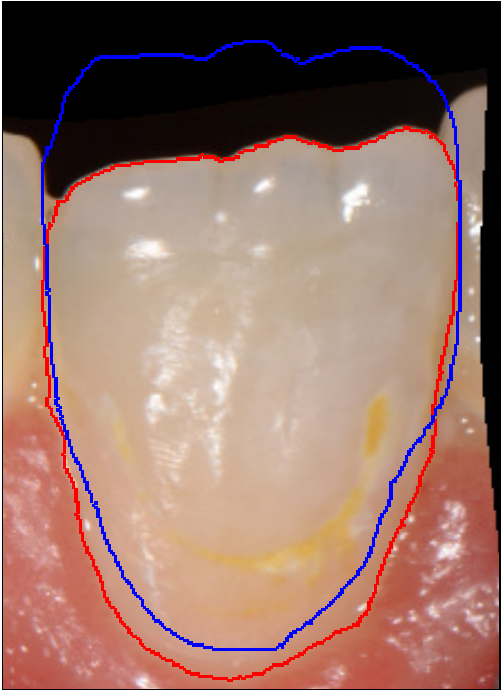}&\includegraphics[height=4.8cm,valign=c]{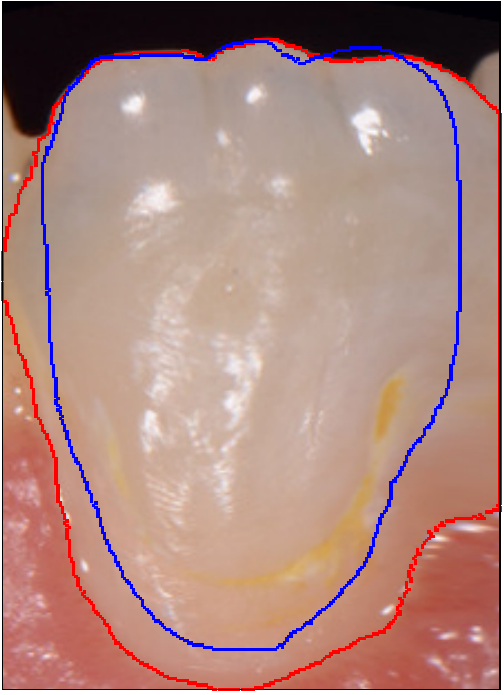}& \makecell{\Large{DC (0.9570)} \\ \Large{HD (0.0571)}} &\makecell{\Large{DC (0.8580)} \\ \Large{HD (0.1600)}}& \makecell{  \hspace{37pt}  \Large{DC (0.8750)} \\ \Large{Worst HD (0.1530)}} \\
\midrule
\bottomrule
\end{tabular}}
\caption{\label{fig:CIR_MI} Best and worst DC and HD for the proposed method (M2R), the ITK mutual information (ITK-MI) and the registration implemented with Elastix. From left to right: QLF with marked GT contour (red), DP with marked GT contour (white), DP with marked GT contour (blue) deformed using the M2R and contour extracted from QLF (red), DP with marked GT contour (blue) deformed using ITK-MI and contour extracted from QLF (red), and DP with marked GT contour (blue) deformed using Elastix and contour extracted from QLF (red).}
\end{figure}

The best and worst DCs and HDs for our method, the ITK-MI registration and the MI-based registration implemented with Elastix are depicted in Fig. \ref{fig:CIR_MI}. The MI-registration implemented with Elastix yields better results than the ITK-MI registration. However, in some cases, the registration implemented with Elastix returned unrealistic deformations, as shown for example in the last row of Figure \ref{fig:CIR_MI}. These results clearly illustrate the improved accuracy of the proposed method over both the ITK-MI and Elastix that was shown quantitatively but in an abstract manner by the DC and HD means.
Disregarding that we have used a non-optimal ROI extractor, we have been able to outperform both the ITK-MI registration and the MI-registration implemented with Elastix significantly. This is due to the ROI-based vs.\ global approach, respectively. It emphasizes the need of ROI-based registration, which is in line with Yi et al.\cite{yi2006}.

Using an optimal ROI segmentation improves the accuracy of the registration with the proposed algorithm. In the case of the DP/QLF alignment, we performed also the registration using the segmentation of the ROIs provided from the ground truth estimated from the manual markings instead of automatically computing the segmentation. This eliminates the segmentation error that is implicitly included in the results shown above. We calculated also the Dice coefficients and the Hausdorff measure for all 150 image pairs. The mean and standard deviation of the DC for the M2R registration with optimal segmentation (ground truth estimated from manual segmentation) are 0.9913 and 0.0018 respectively. The mean and the standard deviation for the HD are 0.0126 and 0.0048. The registration results are therefore even more accurate if an optimal segmentation of the ROI (or classification of the images to be registered into ROI and background) is available.

\begin{figure}[htb]
\centering
\resizebox{\linewidth}{!}{
\begin{tabular}{ccccc}
\begin{tikzpicture}\node at (1.9,2.5){\includegraphics[height = 5.6cm]{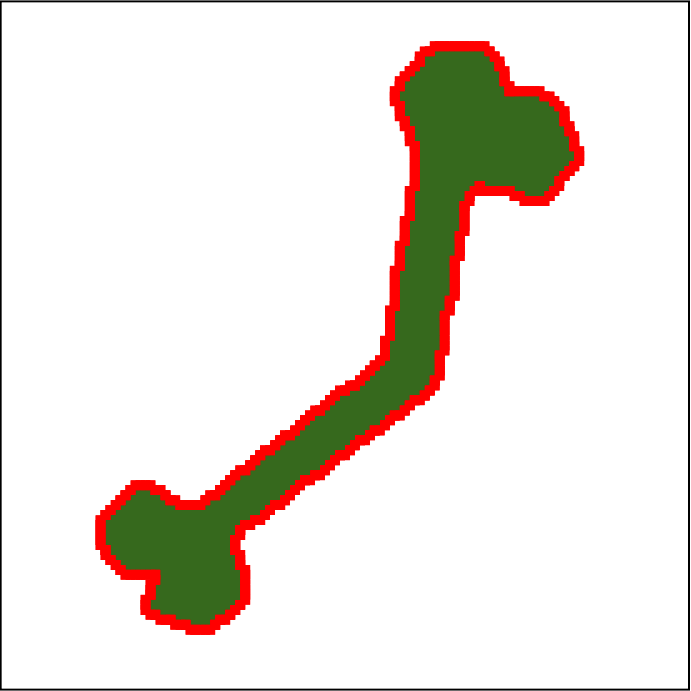}};\node[fill=black,circle,inner sep = 0.9ex] at (-0.4,0.2) { \bf \Large\textcolor{white}{a}};\end{tikzpicture}&\begin{tikzpicture}\node at (3.3,2.5){\includegraphics[height = 5.6cm]{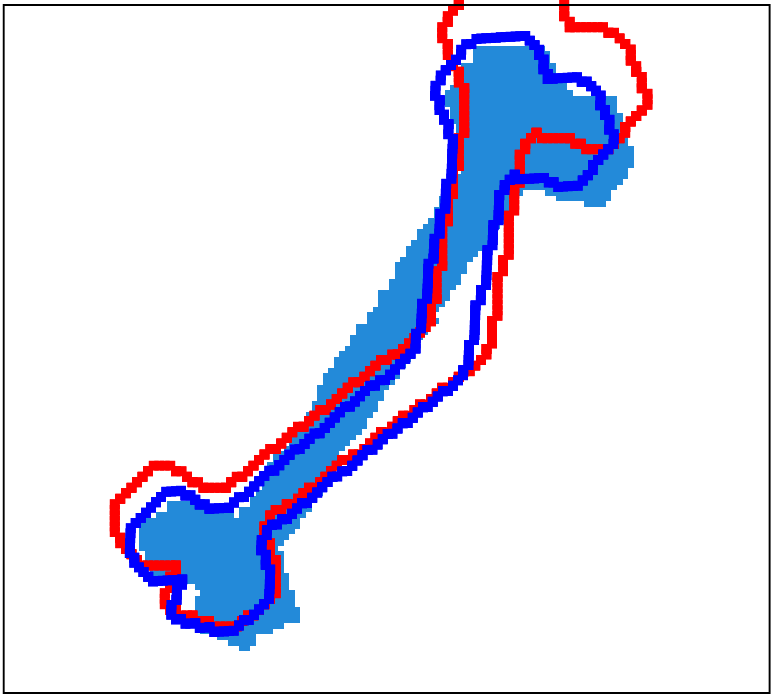}};\node[fill=black,circle,inner sep = 0.9ex] at (.7,0.2) { \bf \Large\textcolor{white}{b}};\end{tikzpicture}&\begin{tikzpicture}\node at (4.3,2.5){\includegraphics[height = 5.6cm]{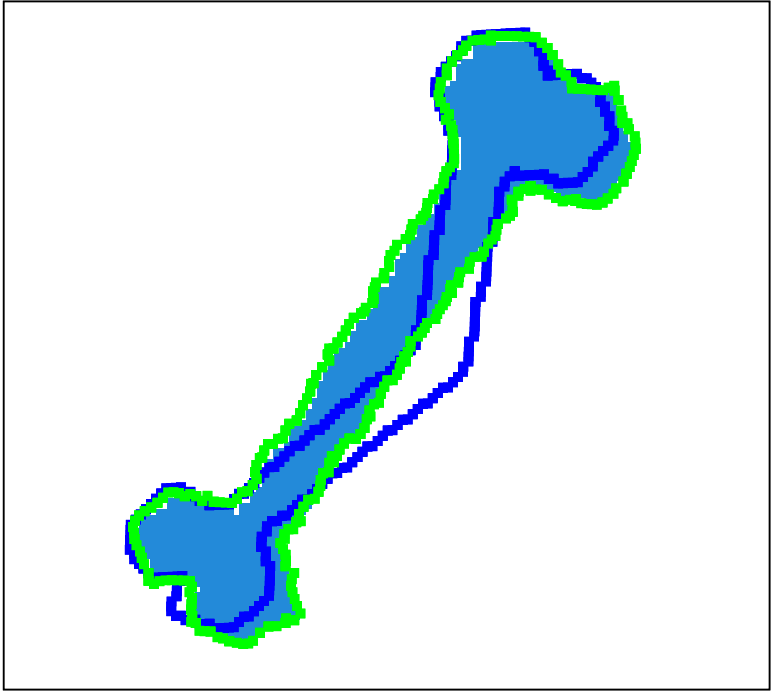}};\node[fill=black,circle,inner sep = 0.9ex] at (1.7,0.2) { \bf \Large\textcolor{white}{c}};\end{tikzpicture}&\begin{tikzpicture}\node at (5.3,2.5){\includegraphics[height = 5.6cm]{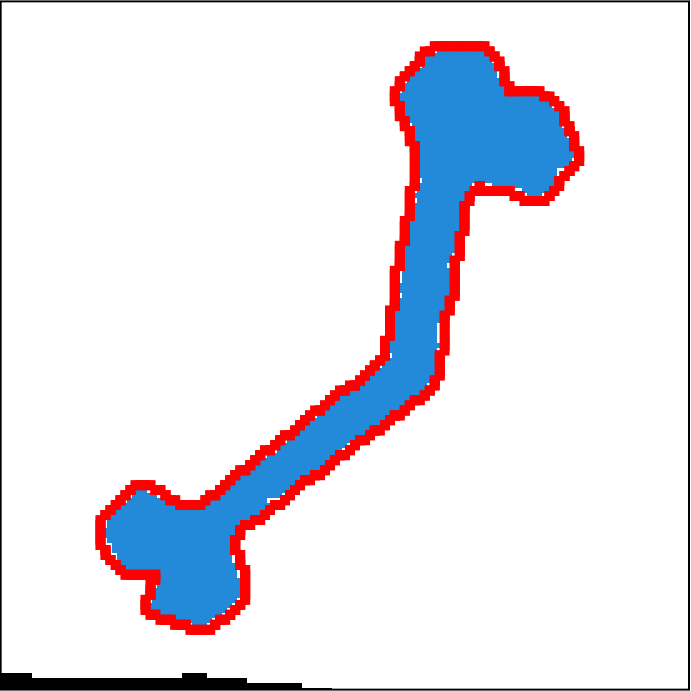}};\node[fill=black,circle,inner sep = 0.7ex] at (3,0.2) { \bf\Large \textcolor{white}{d}};\end{tikzpicture}&\begin{tikzpicture}\node at (6.3,2.5){\includegraphics[height = 5.6cm]{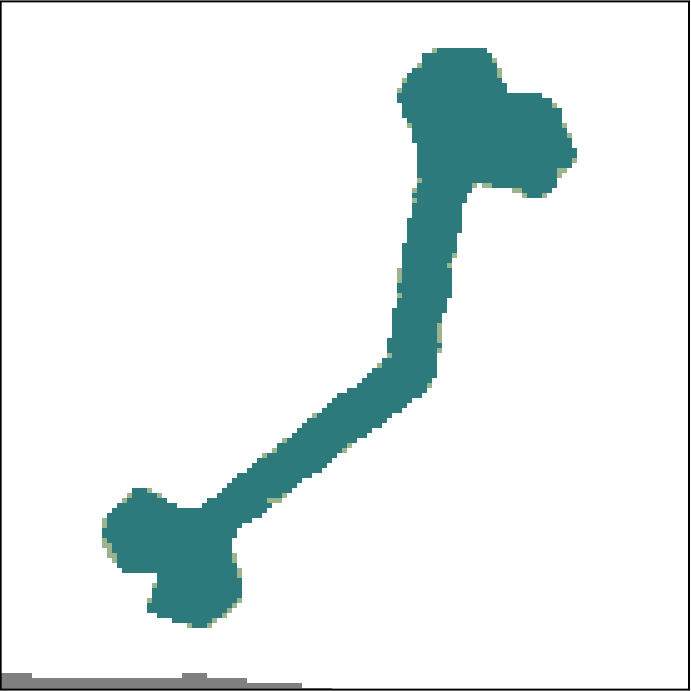}};\node[fill=black,circle,inner sep = 0.9ex] at (4,0.2) { \bf\Large \textcolor{white}{e}};\end{tikzpicture}\\
\begin{tikzpicture}\node at (1.9,2.5){\includegraphics[height = 5.6cm]{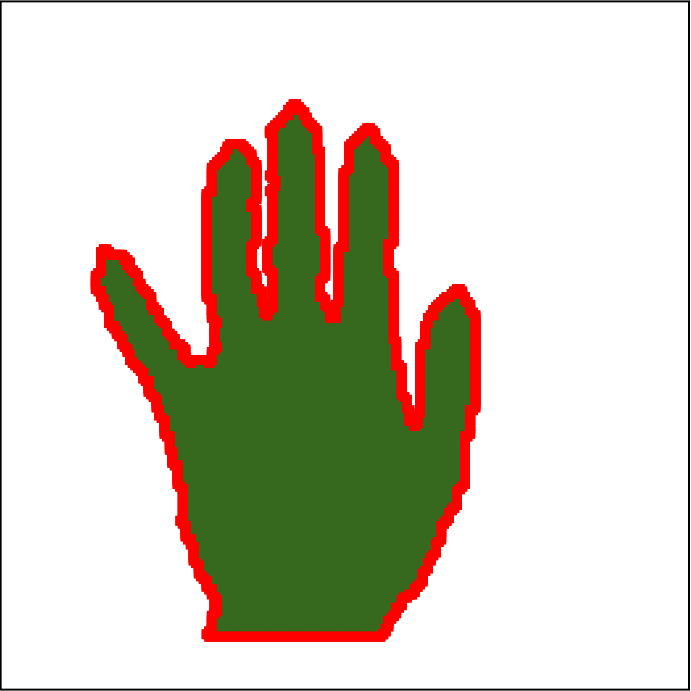}};\node[fill=black,circle,inner sep = 0.9ex] at (-0.4,0.2) { \bf\Large \textcolor{white}{a}};\end{tikzpicture}&\begin{tikzpicture}\node at (3.3,2.5){\includegraphics[height = 5.6cm]{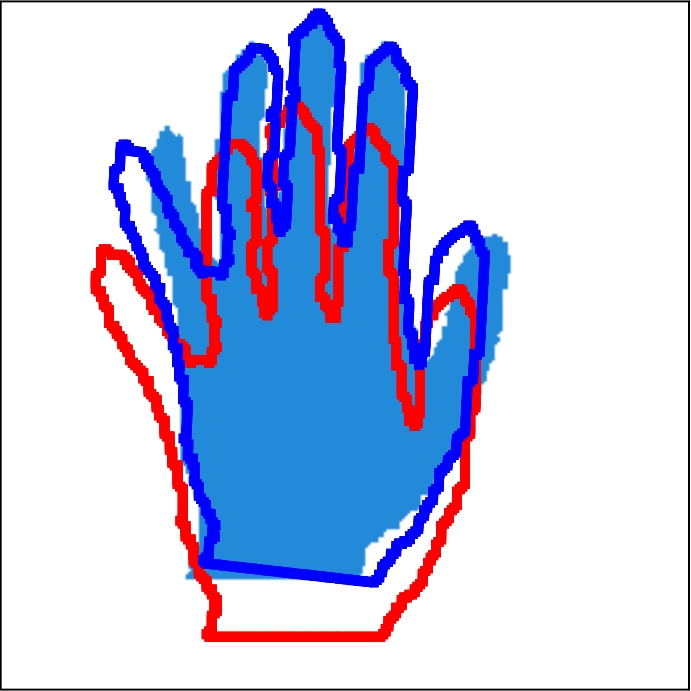}};\node[fill=black,circle,inner sep = 0.9ex] at (1.,0.2) { \bf \Large\textcolor{white}{b}};\end{tikzpicture}&\begin{tikzpicture}\node at (4.3,2.5){\includegraphics[height = 5.6cm]{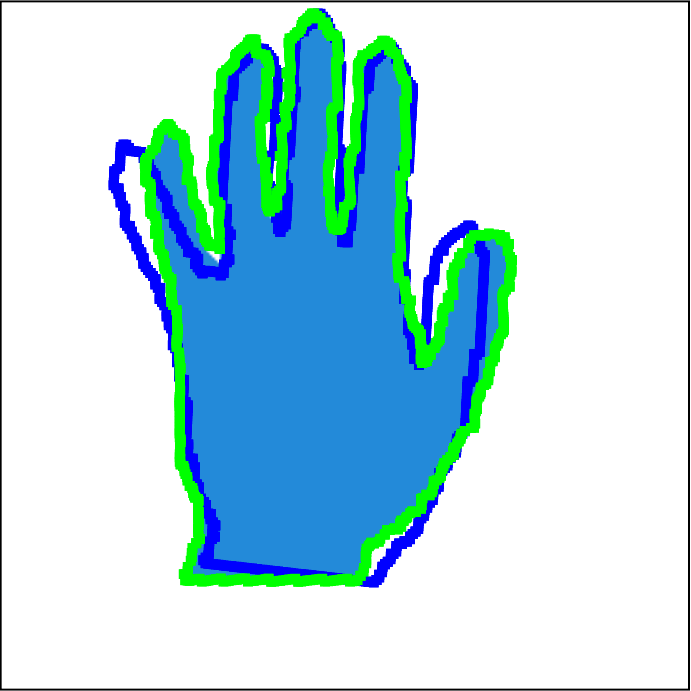}};\node[fill=black,circle,inner sep = 0.9ex] at (2,0.2) { \bf\Large \textcolor{white}{c}};\end{tikzpicture}&\begin{tikzpicture}\node at (5.3,2.5){\includegraphics[height = 5.6cm]{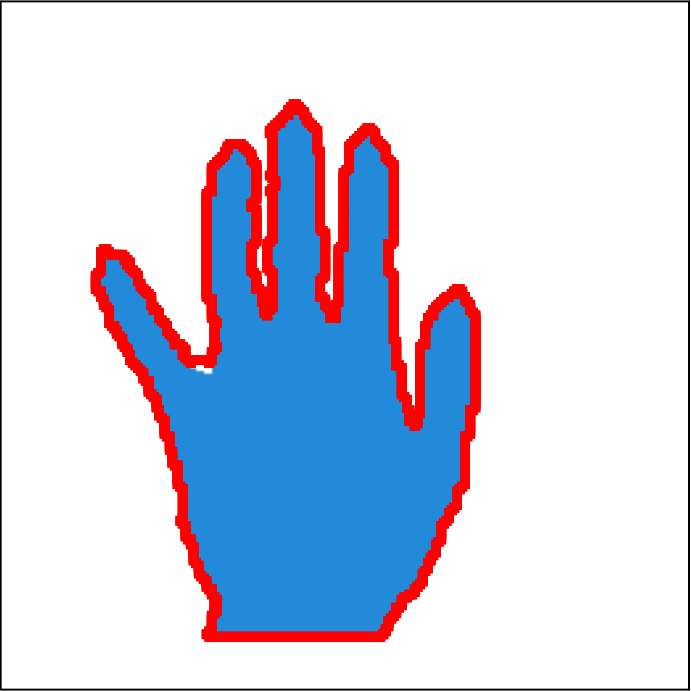}};\node[fill=black,circle,inner sep = 0.9ex] at (3,0.2) { \bf \Large\textcolor{white}{d}};\end{tikzpicture}&\begin{tikzpicture}\node at (6.3,2.5){\includegraphics[height = 5.6cm]{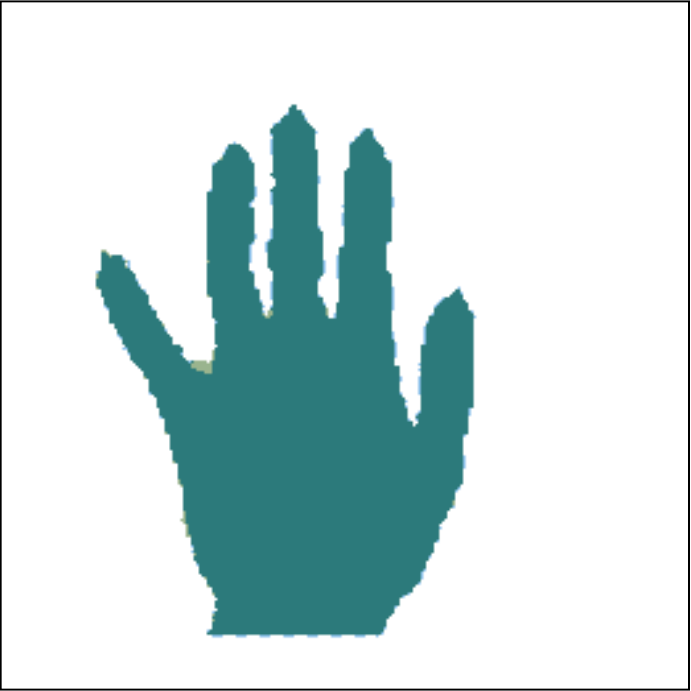}};\node[fill=black,circle,inner sep = 0.9ex] at (4,0.2) { \bf\Large \textcolor{white}{e}};\end{tikzpicture}\\
\begin{tikzpicture}\node at (1.9,2.5){\includegraphics[height = 5.6cm]{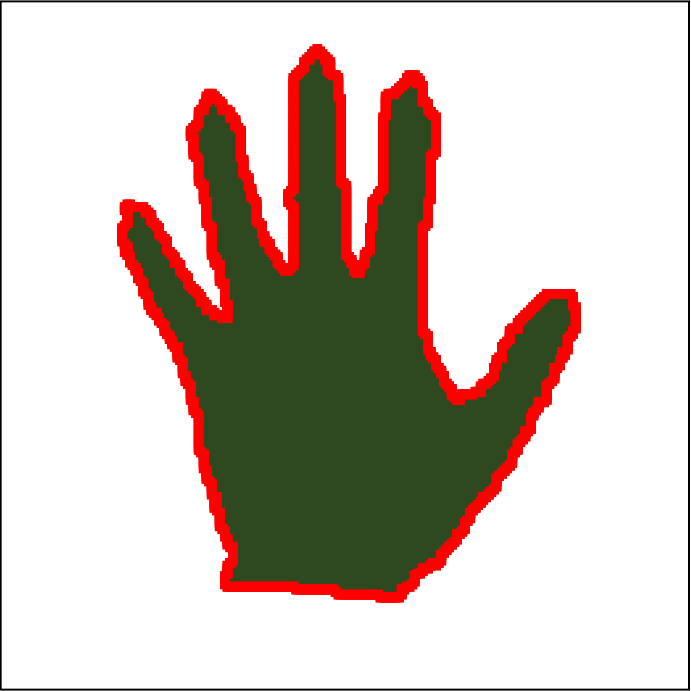}};\node[fill=black,circle,inner sep = 0.9ex] at (-0.4,0.2) { \bf \Large\textcolor{white}{a}};\end{tikzpicture}&\begin{tikzpicture}\node at (3.3,2.5){\includegraphics[height = 5.6cm]{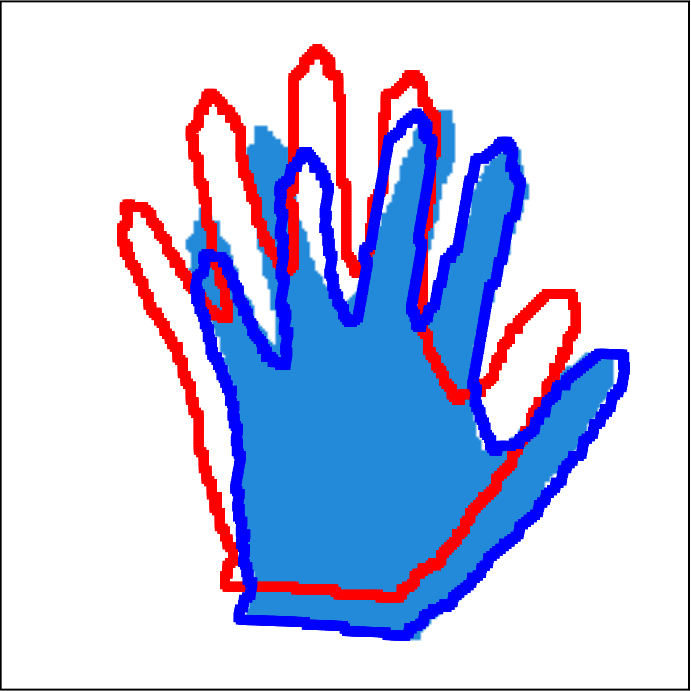}};\node[fill=black,circle,inner sep = 0.9ex] at (1.,0.2) { \bf \Large\textcolor{white}{b}};\end{tikzpicture}&\begin{tikzpicture}\node at (4.3,2.5){\includegraphics[height = 5.6cm]{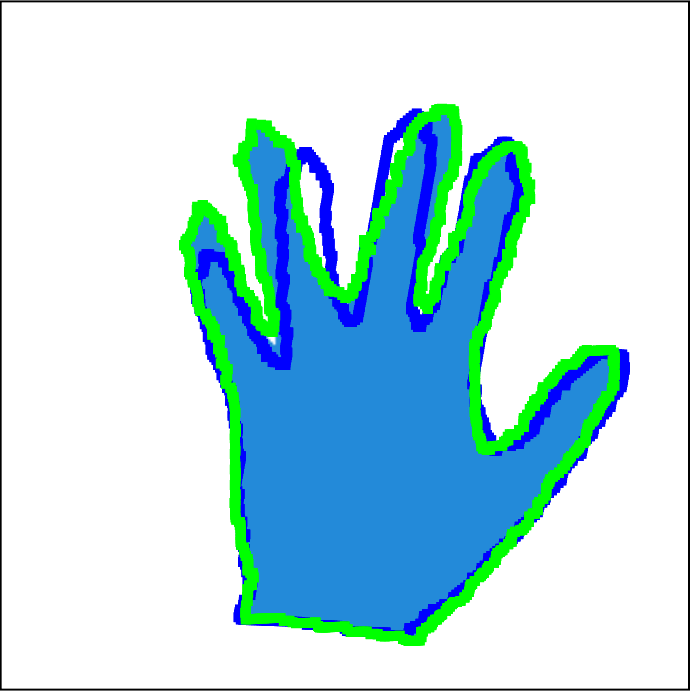}};\node[fill=black,circle,inner sep = 0.9ex] at (2,0.2) { \bf\Large \textcolor{white}{c}};\end{tikzpicture}&\begin{tikzpicture}\node at (5.3,2.5){\includegraphics[height = 5.6cm]{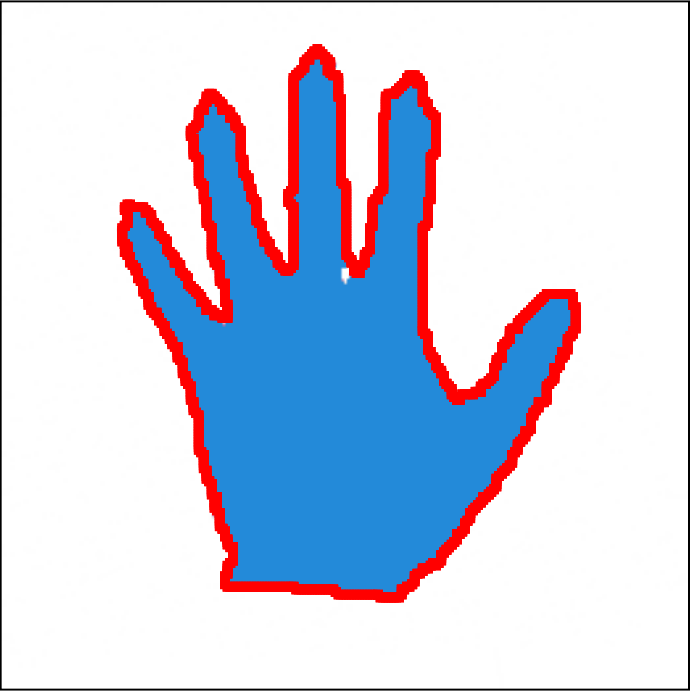}};\node[fill=black,circle,inner sep = 0.9ex] at (3,0.2) { \bf \Large\textcolor{white}{d}};\end{tikzpicture}&\begin{tikzpicture}\node at (6.3,2.5){\includegraphics[height = 5.6cm]{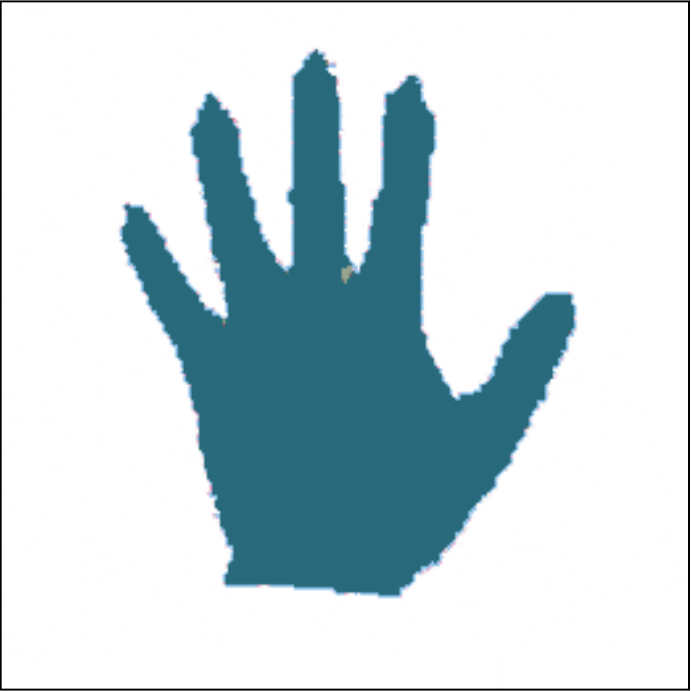}};\node[fill=black,circle,inner sep = 0.9ex] at (4,0.2) { \bf \Large\textcolor{white}{e}};\end{tikzpicture}
\end{tabular}}
\caption{\label{fig:figRes2dShapes} Results for more complex shapes in 2D: The template image with the the superimposed curve (red) (a); the reference with superimposed curves; shape contour before (red) and after parametric registration (blue) (b); and the curves after parametric and non-rigid registration in blue and green, respectively (c). Panel (d) shows the reference after non-rigid deformation and the corresponding template curve (red). Finally, an image created by blending the template and the registered reference is displayed in (e).}
\end{figure}

For both 2D and 3D, the objects of interest presented in our applications are relatively simple in shape. However, the proposed method is capable of handling more complex shapes. As an example in 2D, our method was applied to images of a hand and a bone taken from the 1070-Shape Dataset of the Laboratory for Engineering Man/Machine Systems (LEMS) \cite{LEMS}. The registration results, with values of the parameters $\alpha=1$, $\mu=1$ and $\lambda=10^{-i}$, $i=\{2,3,4,5,6,7\}$, are given in Figure \ref{fig:figRes2dShapes}. The Dice coefficient and the symmetric Hausdorff distance are 0.9819 and 0.0102 for the bone example, 0.9899 and 0.0488 for the first hand example (middle row in Figure \ref{fig:figRes2dShapes}) and 0.9902 and 0.0191 for the second hand example (last row in Figure \ref{fig:figRes2dShapes}).

\begin{figure}[t]
\centering
\resizebox{\linewidth}{!}{
\begin{tabular}{l | l |  l}
\includegraphics[height = 5.7cm]{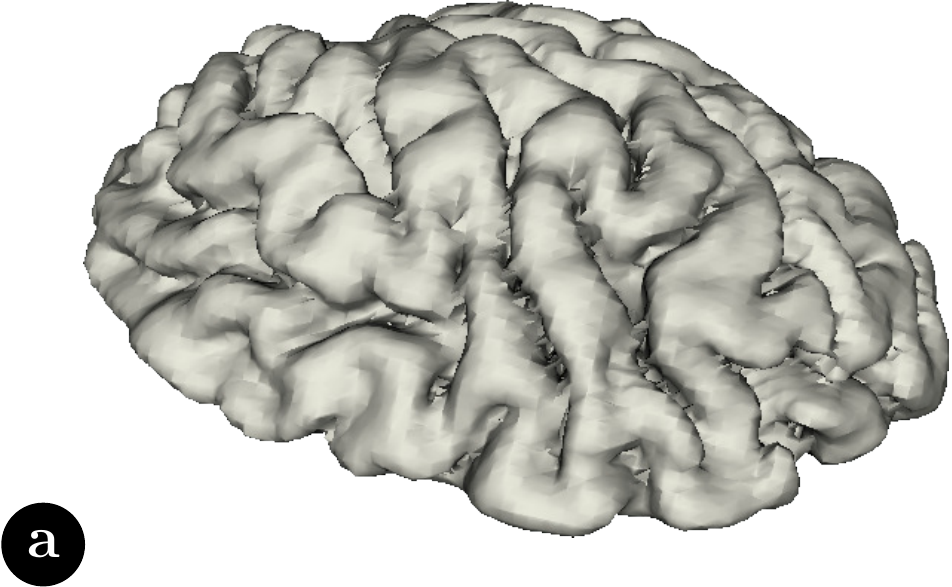}&\includegraphics[height = 5.6cm]{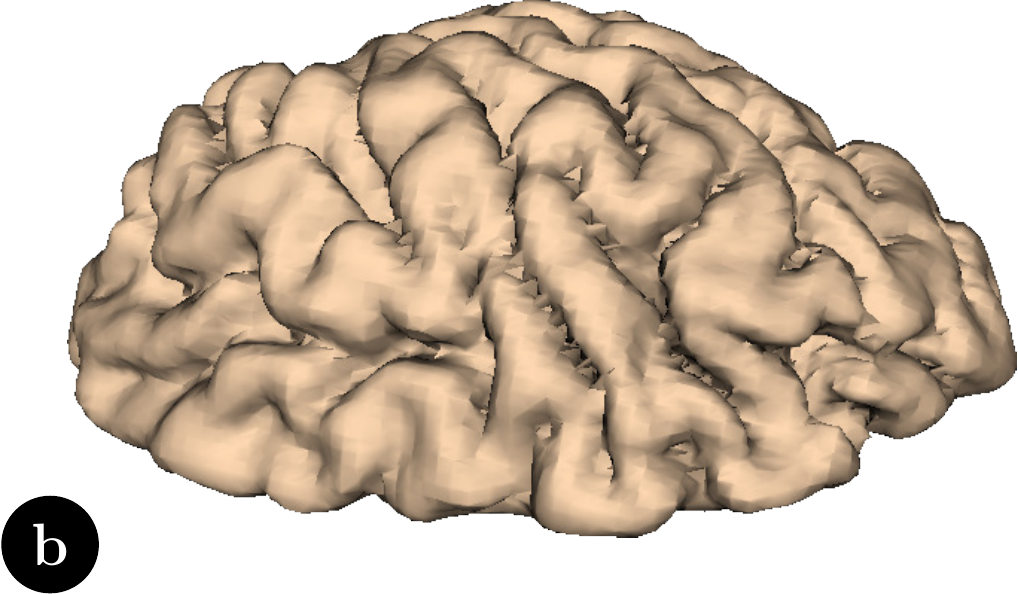}&\includegraphics[height = 5.7cm]{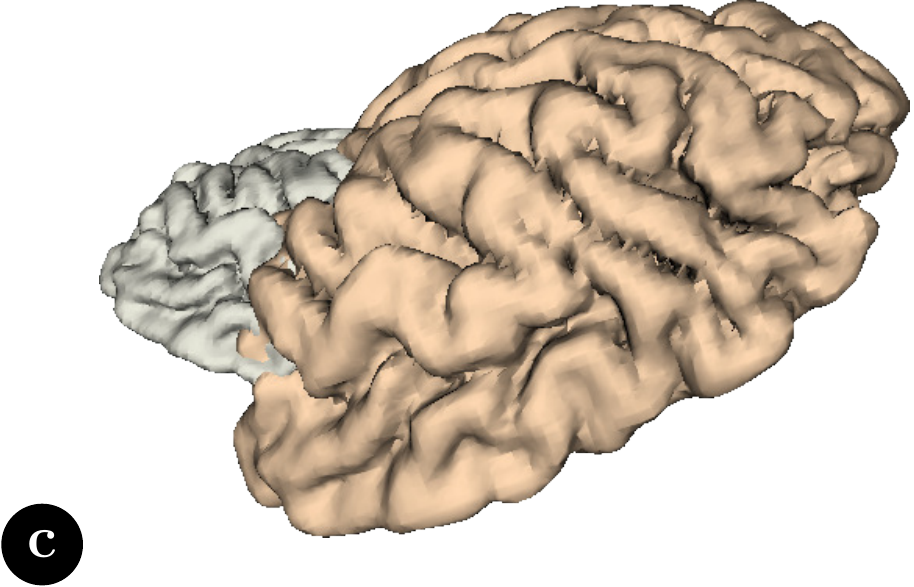}\\
\\ \hline\\
\includegraphics[height = 5.7cm]{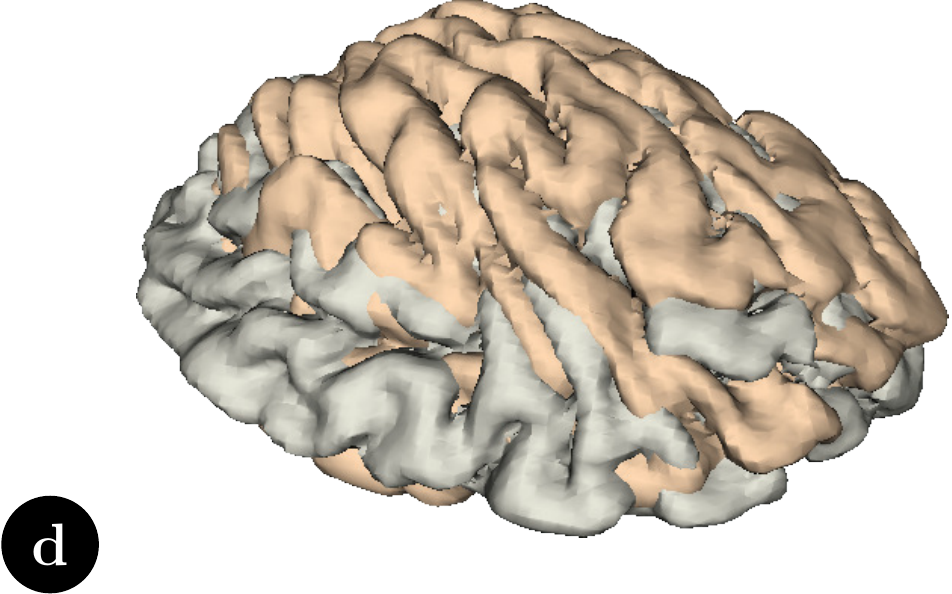}&\includegraphics[height = 5.7cm]{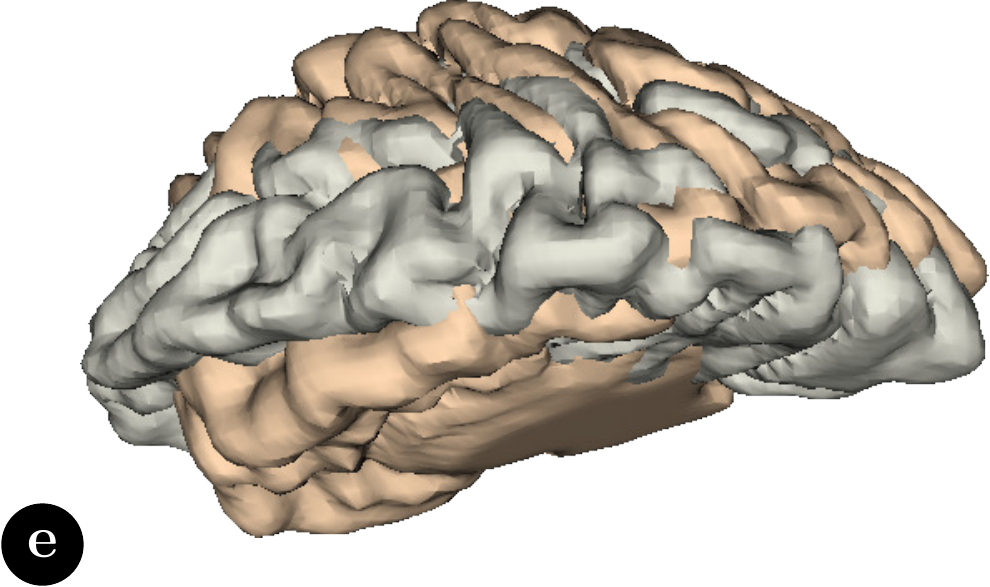}&\includegraphics[height = 5.7cm]{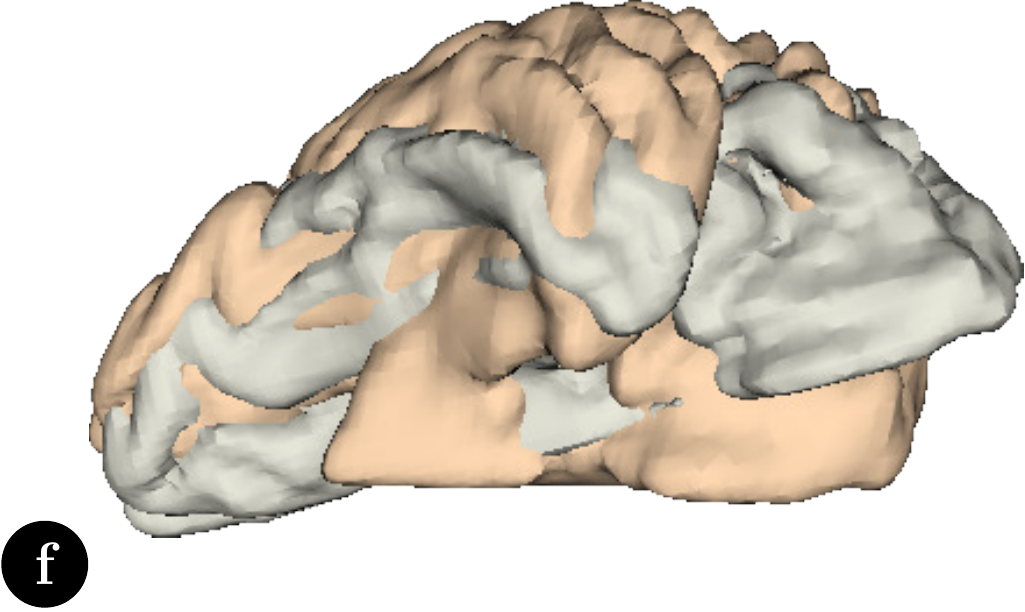}\\
\\ \hline\\
\includegraphics[height = 5.7cm]{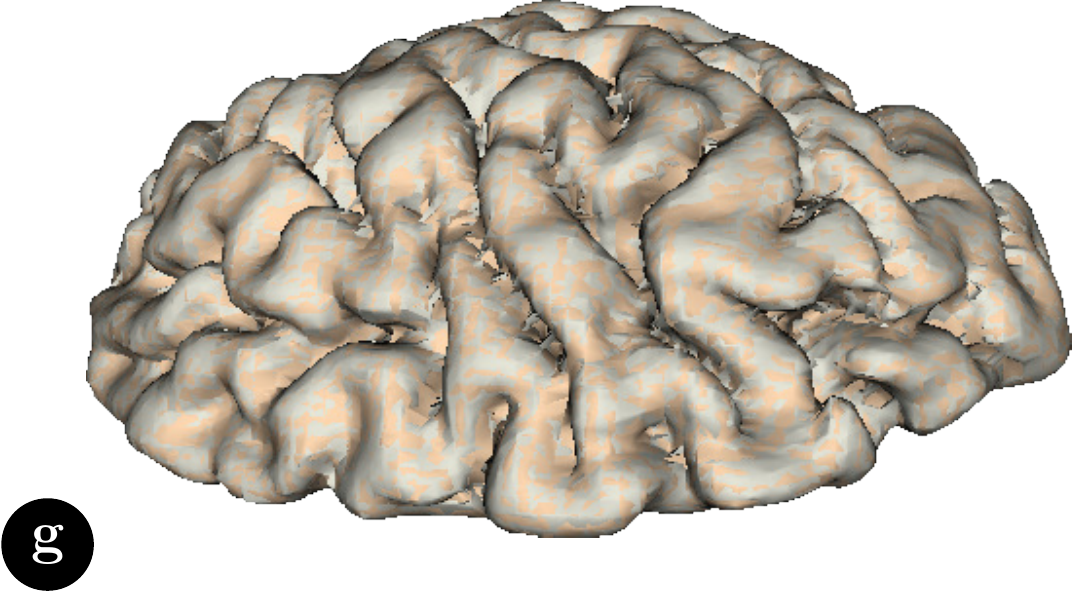}&\includegraphics[height = 5.7cm]{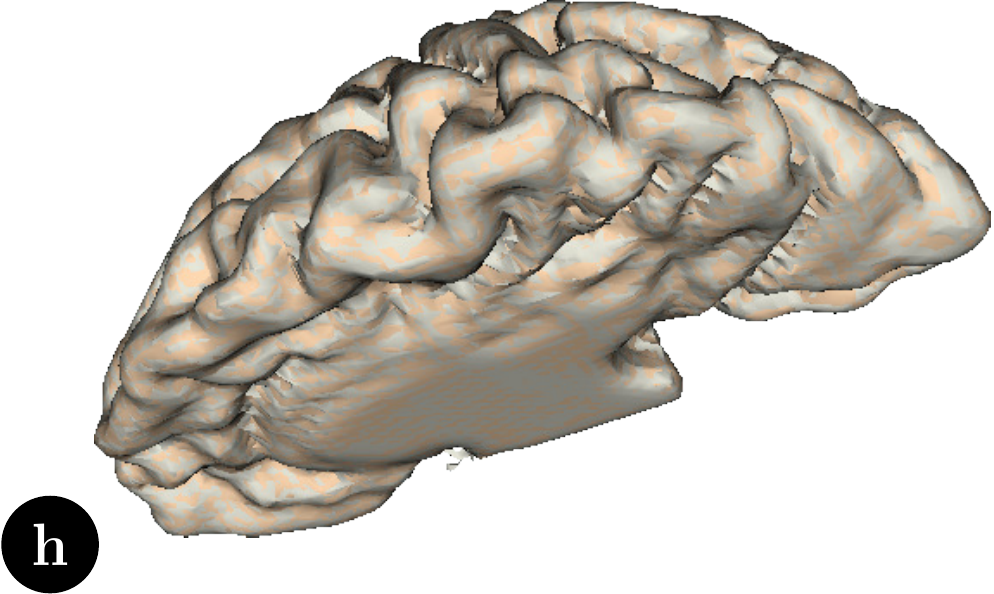}&\includegraphics[height = 5.7cm]{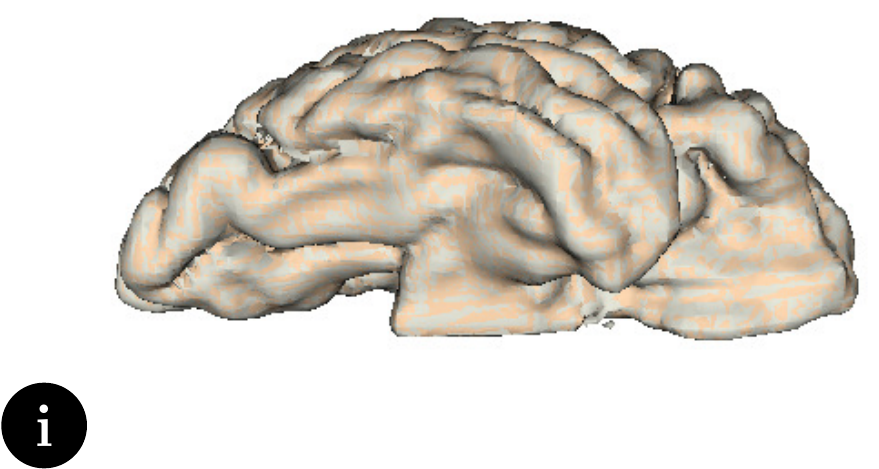}\\
\end{tabular}}
\caption{\label{fig:results3Dbrain} Results in 3D for the brain example.
The first row shows the starting setting: $S_f$ (a), template mesh $g$ (b), and initial position of the input images (c). Note that (a) and (b) use different manually chosen view angles to simplify the comparison, (c) shows the true initial mismatch of the data sets. Middle and lower row depict the results after parametric registration and non-rigid deformation, respectively, from different viewing angles.}
\end{figure}

\begin{figure}[b]
\centering
\resizebox{\linewidth}{!}{
\begin{tabular}{cc|c|c|c|c}
\Huge\bf$\lambda=10$&\includegraphics[height = 5.3cm,valign=c]{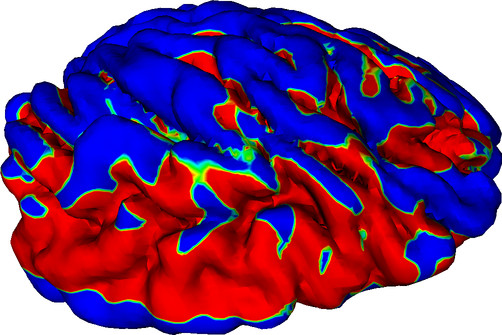}&\includegraphics[height = 5.3cm,valign=c]{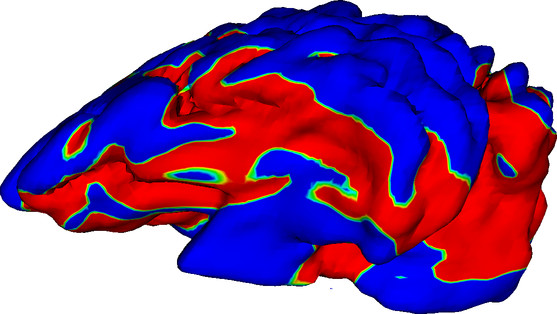}&\includegraphics[height = 5.3cm,valign=c]{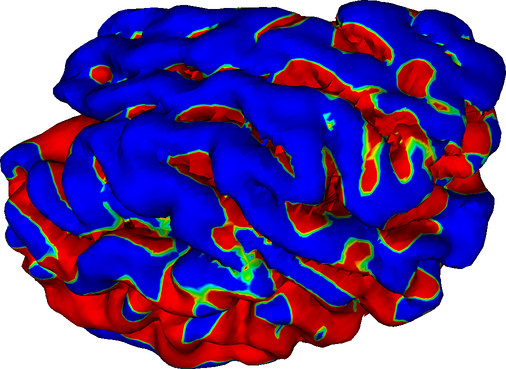}&\includegraphics[height = 5.3cm,valign=c]{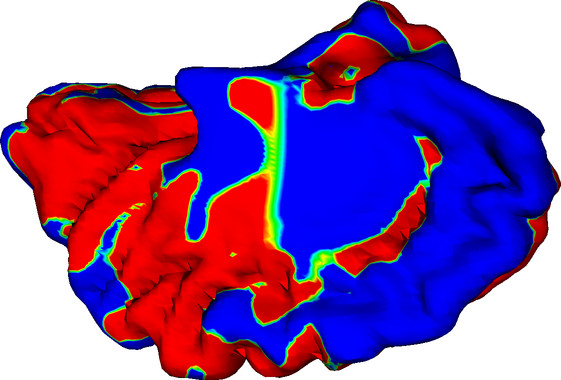}&\includegraphics[height = 5.3cm,valign=c]{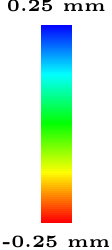}\\
\\ \hline\\
\Huge\bf$\lambda=1$&\includegraphics[height = 5.3cm,valign=c]{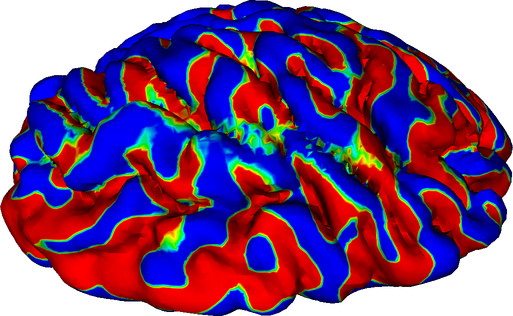}&\includegraphics[height = 5.3cm,valign=c]{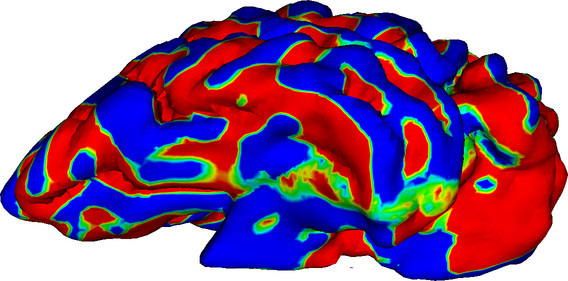}&\includegraphics[height = 5.3cm,valign=c]{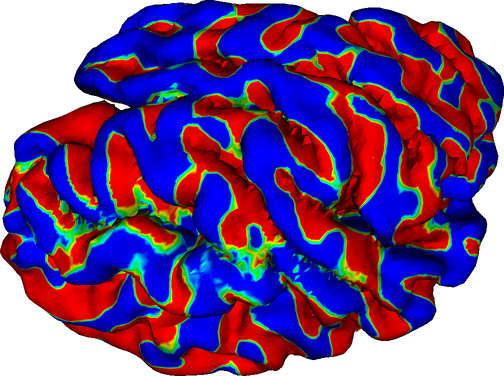}&\includegraphics[height = 5.3cm,valign=c]{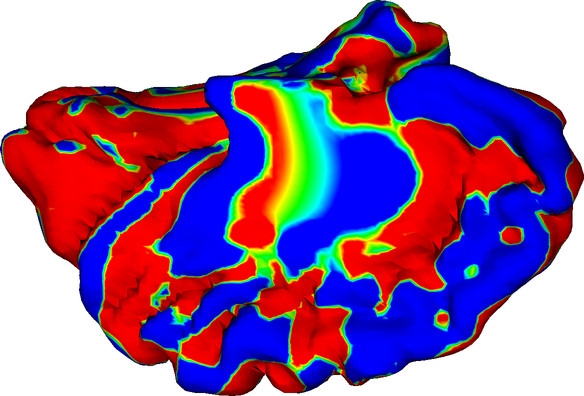}&\includegraphics[height = 5.3cm,valign=c]{colorBarBrain.pdf}\\
\\ \hline\\
\Huge\bf$\lambda=10^{-1}$&\includegraphics[height = 5.3cm,valign=c]{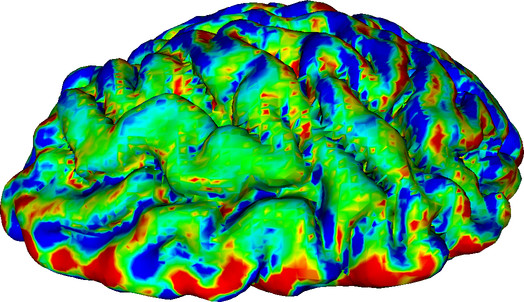}&\includegraphics[height = 5.3cm,valign=c]{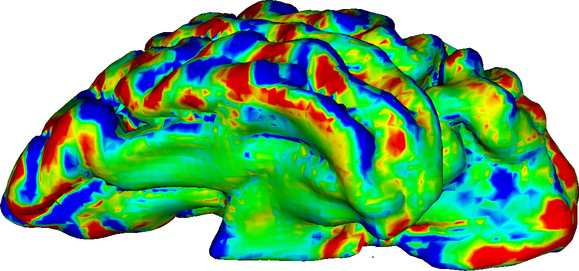}&\includegraphics[height = 5.3cm,valign=c]{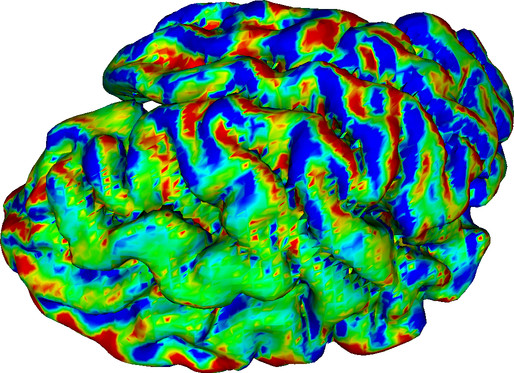}&\includegraphics[height = 5.3cm,valign=c]{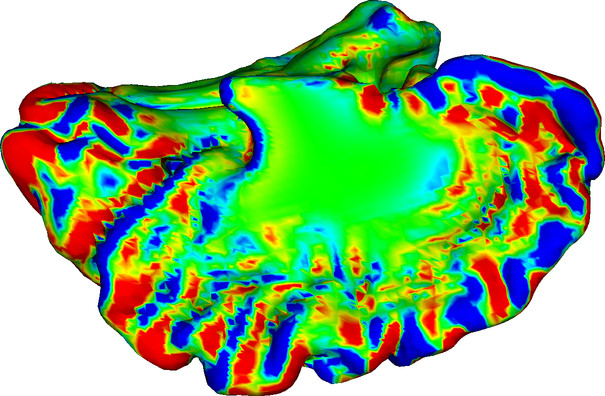}&\includegraphics[height = 5.3cm,valign=c]{colorBarBrain.pdf}\\
\\ \hline\\
\Huge\bf$\lambda=10^{-2}$&\includegraphics[height = 5.3cm,valign=c]{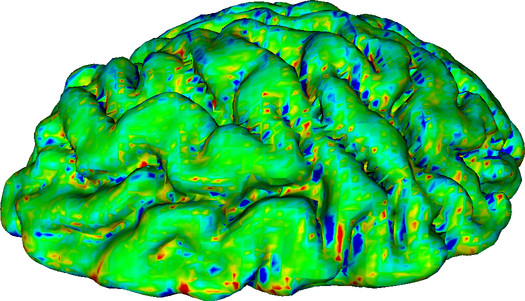}&\includegraphics[height = 5.3cm,valign=c]{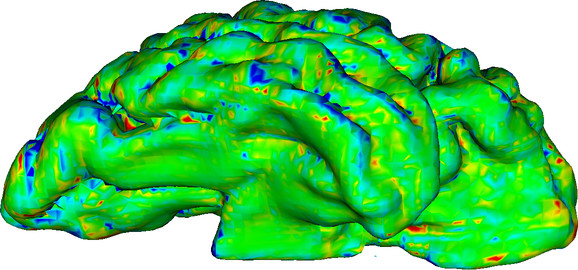}&\includegraphics[height = 5.3cm,valign=c]{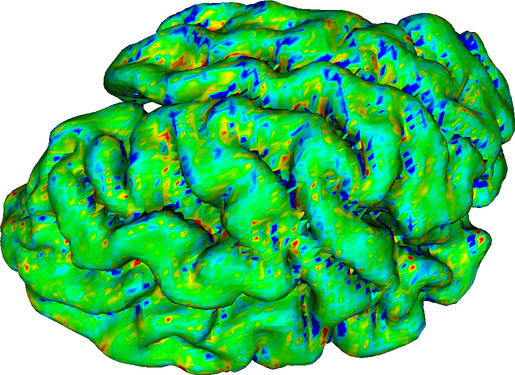}&\includegraphics[height = 5.3cm,valign=c]{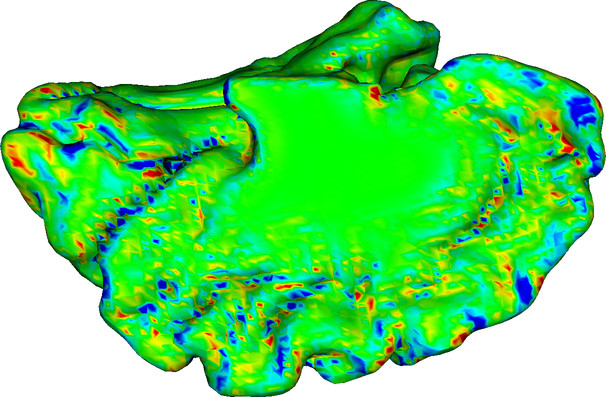}&\includegraphics[height = 5.3cm,valign=c]{colorBarBrain.pdf}\\
\\ \hline\\
\Huge\bf$\lambda=10^{-3}$&\includegraphics[height = 5.3cm,valign=c]{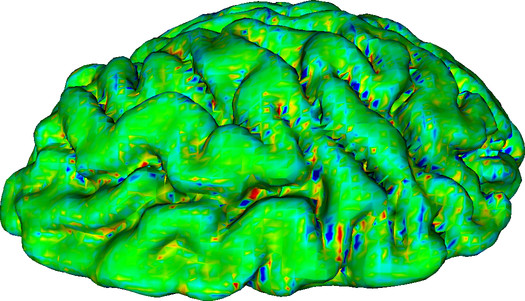}&\includegraphics[height = 5.3cm,valign=c]{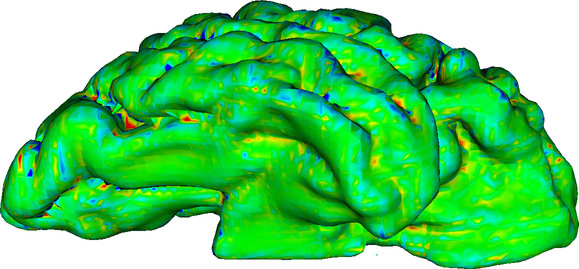}&\includegraphics[height = 5.3cm,valign=c]{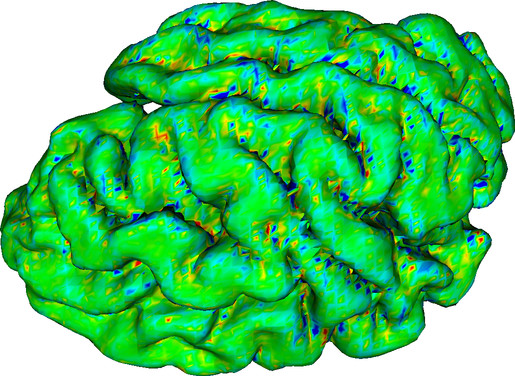}&\includegraphics[height = 5.3cm,valign=c]{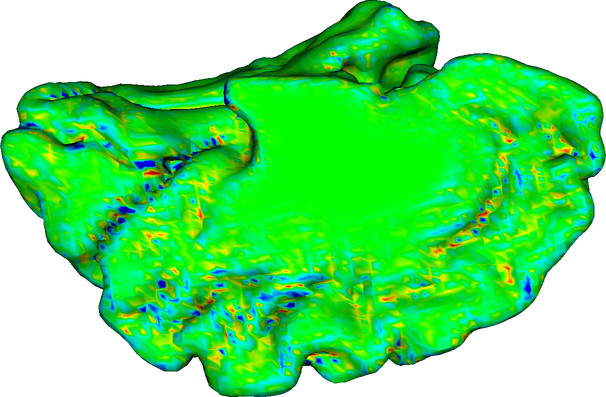}&\includegraphics[height = 5.3cm,valign=c]{colorBarBrain.pdf}\\
\\ \hline\\
\Huge\bf$\lambda=10^{-4}$&\includegraphics[height = 5.3cm,valign=c]{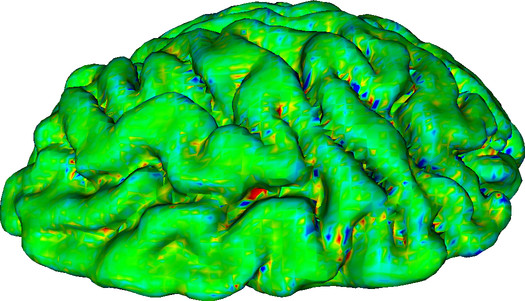}&\includegraphics[height = 5.3cm,valign=c]{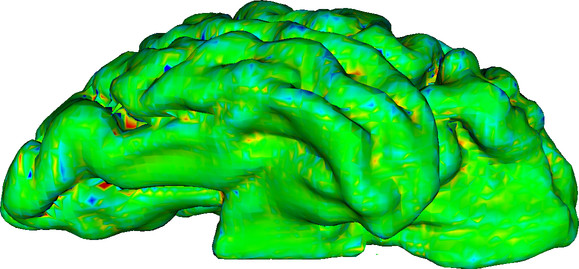}&\includegraphics[height = 5.3cm,valign=c]{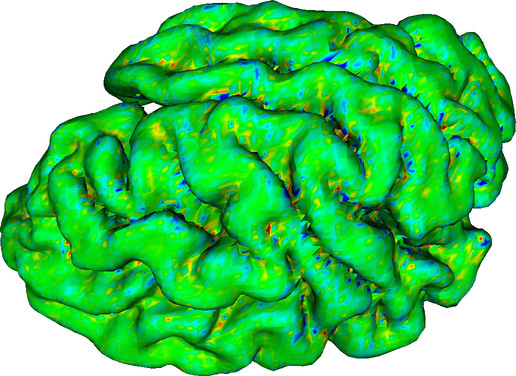}&\includegraphics[height = 5.3cm,valign=c]{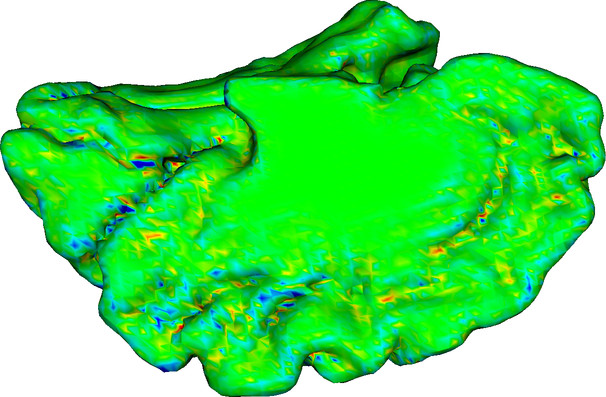}&\includegraphics[height = 5.3cm,valign=c]{colorBarBrain.pdf}\\
\\ \hline\\
\Huge\bf$\lambda=10^{-5}$&\includegraphics[height = 5.3cm,valign=c]{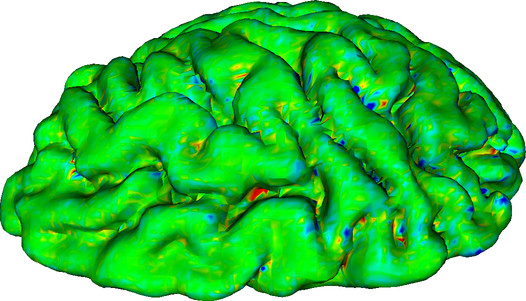}&\includegraphics[height = 5.3cm,valign=c]{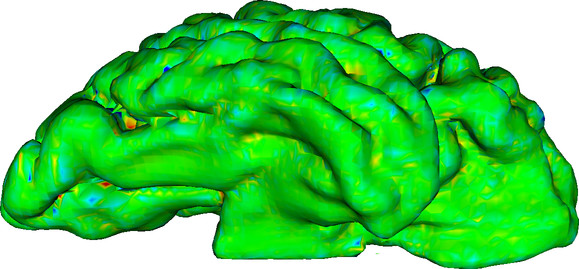}&\includegraphics[height = 5.3cm,valign=c]{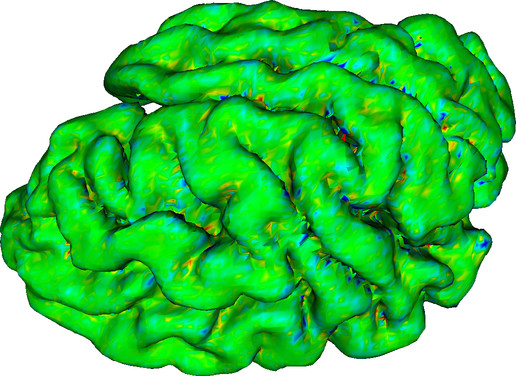}&\includegraphics[height = 5.3cm,valign=c]{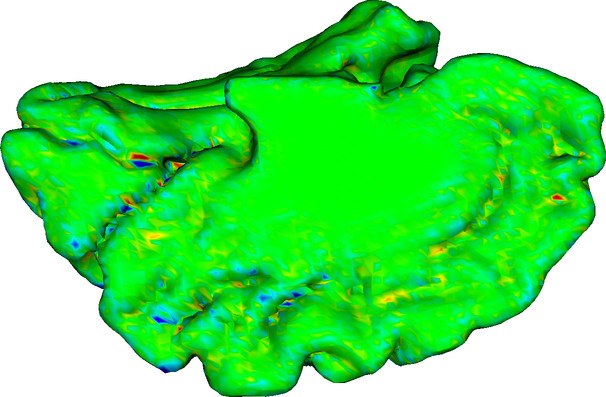}&\includegraphics[height = 5.3cm,valign=c]{colorBarBrain.pdf}\\
\end{tabular}}
\caption{\label{fig:results3DcoloredBrain} The distances $d(\phi(c), S_f)$, for $c\in\mathcal{C}$, are displayed as color coding on the registered template mesh $\phi(\mathcal{C})$ for every value of the parameter $\lambda$ used for the non-rigid registration. From left to right: $\phi(\mathcal{C})$ as seen from four different angles, color bar. The measure of the distance in mm was estimated considering an average acquisition field of view for head CT scans of 250 mm. }
\end{figure}

As an example of application of the proposed method to more complicated shapes in 3D, a brain CT scan is used. The template has been obtained by applying a synthetic deformation to the reference mesh. The considered deformation was defined as $u_1(x) = -\beta x_1(x_2-\frac{1}{2})\allowbreak+\allowbreak\beta (1-x_1)(x_1-\frac{1}{2})\allowbreak+\beta (1-x_1)(x_3-\frac{1}{2})$, $u_2(x)=\beta x_1(x_2-\frac{1}{2})+\beta(1-x_1)(x_2-\frac{1}{2})-\beta(1-x_1)(x_3-\frac{1}{2})$, $u_3(x)=\allowbreak  -\beta x_1(x_3-\frac{1}{2})\allowbreak +\allowbreak  \beta(1-x_1)(x_3-\frac{1}{2})$, with $\beta=0.3$. After applying the defined nonlinear deformation to the reference mesh, a parametric rigid body transformation was applied to the resulting mesh. The chosen parameters were $t=(t_1,t_2,t_3)=(0.,0.,0.1)$ as translation vector and $\theta=(\theta_1, \theta_2,\theta_3)=(0.1,\allowbreak  0.5,-0.03)$ radians for the rotation angles. Figure
 \ref{fig:results3Dbrain} shows the results of the proposed registration method for the brain example, using $\alpha=10^2$, $\mu=10^2$ and $\lambda=10^{-i},\;i=\{-1,0,1,2,3,4,5\}$ as parameters values. As in the hips example, the resolution of the  3D grid is $129\times 129\times129$, with pixels dimensions (1.9380, 1.9380, 1.9380) mm.  Figure \ref{fig:results3DcoloredBrain} displays the distances $d(\phi(c), S_f)$.
The Dice coefficient after the registration is 0.9993. The two one-sided Hausdorff distances before the registration are 100.72 mm (51.97 pixel units) for the reference mesh from the template and 45.30 mm (23.38 pixel units) for the other direction. The 99th percentiles of the Hausdorff distance of the reference mesh from the template are 94.82 mm (48.93 pixel units) and 39.50 mm (20.38 pixel units) for the other direction. After the registration, the one-sided Hausdorff distance of the reference mesh from the registered template is 1.35 mm (0.0054 pixel units) and 6.27 mm (0.0251) for the other direction, while the 99th percentiles are 1 mm (0.52 pixel units) and 2.65 mm (1.37 pixel units) for the Hausdorff distance of the reference mesh from the registered template and the other direction respectively. The noticeable difference between the Hausdorff distance and its 99th percentile is due to the presence of a few outliers, displayed as colored (red or blue) dots in Figure \ref{fig:results3DcoloredBrain}.

A limitation of our work is that the extensive quantitative evaluation is performed only in 2D. In future, we plan to obtain extensive quantitative assessment in 3D, too. Here data from the Evaluation of Methods for Pulmonary Image Registration 2010 (EMPIRE10) or other challenges might be helpful \cite{murphy2011}. In this context, the proposed method could be used to perform lung registration by aligning the lung boundaries. However, to align the major fissures inside the lung, an additional data term would need to be included in the proposed method, which we plan to exploit in a future work.

Since the proposed method can also be used for segmentation, even though it is performed as a pre-processing step, it would be interesting to explore the possibility of extending the method to an integrated segmentation/registration scheme in the future.

\section{Conclusions}
A ROI-based registration method for multi-modal images was presented. It uses a curve-to-pixel or surface-to-voxel approach to align the ROIs from the reference and the target images in 2D or 3D, respectively. Qualitative examples in 2D and 3D were presented. The accuracy of the alignment was tested on multi-modal 2D images, as well as in 3D examples, by calculating the Dice coefficient and the Hausdorff distance between the registered images. For the quantitative evaluation in 2D, the ground truth was established by applying the STAPLE algorithm to manually marked images. In a comprehensive analysis based on 150 pairs of images, the proposed method statistical significantly outperforms mutual information (MI)-based global registration implemented using both ITK and Elastix, which is considered as state-of-the-art method.

\newpage
\appendix
\section{details of the parametric registration in 2D}\label{app-1}
For the parametric registration in 2D, if $A=\left(\begin{matrix}a_{11} & a_{12}\\ a_{21} & a_{22}\end{matrix} \right)$, $D_A=\left(\begin{matrix}a_{11} & 0\\ 0 & a_{22}\end{matrix} \right)$, the regularizer can be expressed as $$E_\text{reg}[\varphi] = \frac{\alpha^2}{2}\left((a_{11}-1)^2+a_{12}^2+a_{21}^2+(a_{22}-1)^2\right)+\frac{\mu^2}{2}\left(\left(1-\frac{a_{11}}{a_{22}}\right)^2+\left(1-\frac{a_{22}}{a_{11}}\right)^2\right)_.$$ Therefore, the vector $F$ for the parametric registration is
$$F[\varphi]= \left[\begin{array}{c}\left\{\sqrt{w_{c_i}l_{c_i}}d\left(\varphi\left(c_{i+\frac{1}{2}}\right), S_f\right)\right\}_{i=1,\dots,N}\\ \alpha (1-a_{11})\\ \alpha a_{12}\\ \alpha a_{21}\\ \alpha (1-a_{22})\\ \mu \left(1-\frac{a_{11}}{a_{22}}\right)\\ \mu \left(1-\frac{a_{22}}{a_{11}}\right)\end{array}\right].$$

\section{details of the parametric registration in 3D}\label{app-2}

For the parametric registration in 3D, if $A=\left(\begin{smallmatrix}a_{11} & a_{12} & a_{13} \\ a_{21} & a_{22}& a_{23}\\ a_{31}& a_{32}& a_{33}\end{smallmatrix} \right)$, $D_A=\left(\begin{smallmatrix}a_{11} & 0 &0\\ 0 & a_{22}&0\\ 0&0&a_{33}\end{smallmatrix} \right)$, the regularizer for the parametric registration, (\ref{eq:regPar}), can be expressed as
{\footnotesize{\begin{align*}E_\text{reg}^\text{par}[\varphi] &= \frac{\alpha^2}{2}\left((a_{11}-1)^2+a_{12}^2+a_{13}^2+a_{21}^2+(a_{22}-1)^2+a_{23}^2+a_{31}^2+a_{32}^2+(a_{33}-1)^2\right)\\&+\frac{\mu^2}{2}\left(\left(1-\frac{a_{11}}{a_{22}}\right)^2+\left(1-\frac{a_{11}}{a_{33}}\right)^2+\left(1-\frac{a_{22}}{a_{11}}\right)^2+\left(1-\frac{a_{22}}{a_{33}}\right)^2+\left(1-\frac{a_{33}}{a_{11}}\right)^2+\left(1-\frac{a_{33}}{a_{22}}\right)^2\right).\end{align*}}} Therefore, the vector $F$ for the parametric registration is
$$F[\varphi]= \left[\begin{array}{c}\left\{\sqrt{m_T^q}d(x_T^q+u(x_T^q), S_f)\right\}_{T\in\mathcal{T}}\\ \alpha (1-a_{11})\\ \alpha a_{12}\\ \alpha a_{13}\\ \alpha a_{21}\\ \alpha (1-a_{22})\\ \alpha a_{23}\\ \alpha a_{31}\\ \alpha a_{32}\\ \alpha (1-a_{33})\\ \mu \left(1-\frac{a_{11}}{a_{22}}\right)\\  \mu \left(1-\frac{a_{11}}{a_{33}}\right)\\ \mu \left(1-\frac{a_{22}}{a_{11}}\right)\\  \mu \left(1-\frac{a_{22}}{a_{33}}\right)\\  \mu \left(1-\frac{a_{33}}{a_{11}}\right)\\  \mu \left(1-\frac{a_{33}}{a_{22}}\right)\\ \end{array}\right].$$

\section{details of the ANOVA for the segmentation evaluation}\label{app-3}

This section presents the results ($F$ and $p-$value) of the one-way repeated measures ANOVA for the comparison of the means of the Dice coefficient (DC) as well as of the symmetric Hausdorff distance (HD) of each human rater and the automatic segmentation for both the QLF and the DP.

\begin{table}[H]
\caption{\label{tab:ANOVA} ANOVA results for the comparison of the means of the Dice coefficient (DC) and symmetric Hausdorff distance (HD) of the human raters and the automatic segmentation for QLF and DP.}
\vspace{0.3cm}
\centering
\resizebox{\linewidth}{!}{\begin{tabular}{c|rc|rc|rc|rc}\hline
{\bf Modality} &\multicolumn{4}{c|}{QLF}&\multicolumn{4}{c}{DP}\\
\hline
{\bf Metrics} & \multicolumn{2}{c}{DC} & \multicolumn{2}{c|}{HD}& \multicolumn{2}{c}{DC} & \multicolumn{2}{c}{HD}\\
& $F(1,149)$ & $p$ & $F(1,149)$ & $p$ & $F(1,149)$ & $p$ & $F(1,149)$ & $p$ \\
\hline
{\bf }& &     &        &        &        &        &        &       \\
(R1,M2R)  & 113.16        & $<0.001$ & 107.73 & $<0.001$ & 20.01        & $<0.001$ & 54.85 & $<0.001$\\
(R2,M2R)  & 8.76      & 0.0036 & 38.34 & $<0.001$ & 155.08       & $<0.001$ & 139.09 & $<0.001$\\
(R3,M2R)  & 234.00  & $<0.001$ & 134.41 & $<0.001$ & 415.61 & $<0.001$ & 345.48 & $<0.001$\\
\hline
\end{tabular}}
\end{table}

\section*{Informed consent} Informed consent was obtained from all individuals participating in the trial. The study was approved
by the IRB no IORG0006299 (Ethics Committee, Uniklinik RWTH Aachen).
\section*{Disclosures} No conflicts of interest, financial or otherwise, are declared by the authors.

\acknowledgments
The authors would like to thank Eva E.\ Ehrlich and Ekaterina Sirazitdinova, Uniklinik RWTH Aachen, for their assistance in the preparation of the 2D data. The authors at AICES RWTH Aachen were funded in part by the Excellence Initiative of the German Federal and State Governments.
The RASimAs project has received funding from the European Union's Seventh Framework Programme for research, technological development and demonstration under grant agreement no 610425. The authors declare no conflict of interest and have nothing to disclose.


\bibliography{paperRevision}
\bibliographystyle{spiejour}

\end{document}